\documentclass[11pt]{article}

\usepackage{acl}

\usepackage[normalem]{ulem}
\usepackage{xspace}
\usepackage{graphicx}
\usepackage{xcolor}
\usepackage{color}

\usepackage{times}
\usepackage{caption}
\usepackage{subcaption}

\usepackage{advdate}
\usepackage{datetime}
\usepackage{datenumber}

\usepackage{multirow}   
\usepackage{booktabs}   
\usepackage[flushleft]{threeparttable}
\usepackage{arydshln}       %
\usepackage{makecell}
\usepackage{array}
\usepackage[ruled,vlined]{algorithm2e}

\usepackage[utf8]{inputenc} %
\usepackage[T1]{fontenc}    %
\usepackage{hyperref}       %
\usepackage{url}            %
\usepackage{nicefrac}       %
\usepackage{enumitem}       %
\usepackage{microtype}

\usepackage{soul}           %
\usepackage{color}

\usepackage{lipsum}
\usepackage{fancyvrb}
\usepackage{fvextra}

\usepackage{ifthen}
\usepackage{totcount}

\usepackage{latexsym}
\usepackage{tikz}
\usepackage{amsmath, amssymb, amsthm}
\DeclareMathOperator*{\argmax}{arg\,max}
\usetikzlibrary{bayesnet}
\usetikzlibrary{positioning}
\usetikzlibrary{arrows}
\usepackage[capitalise]{cleveref}
\usepackage{inconsolata}
\usepackage{calc}

\usepackage[inline,nomargin]{fixme}
\fxsetup{theme=color}
\usepackage{bbm}
\newcommand{\ind}{\mathbbm{1}}

\usepackage{pifont}

\newcommand{\daystotarget}[3]{
    \newcounter{targetdate}
    \setmydatenumber{targetdate}{#1}{#2}{#3}
    \newcounter{today}
    \setmydatenumber{today}{\the\year}{\the\month}{\the\day}
    \addtocounter{targetdate}{-\thetoday}
    \thetargetdate\xspace
}

\newtotcounter{taskcounter}
\newtotcounter{taskcounterme}
\newboolean{showtasks}
\setboolean{showtasks}{true}

\newcommand{\task}[1]{
  \addtocounter{taskcounter}{1}
  \ifthenelse{\boolean{showtasks}}
  {{\color{red} [Task: #1]}\xspace}
  {}
}
\newcommand{\taskme}[1]{
  \addtocounter{taskcounterme}{1}
  \ifthenelse{\boolean{showtasks}}
  {{\definecolor{foo}{HTML}{339AF0}\sethlcolor{foo}\color{foo}\emph{#1}}\xspace}
  {}
}

\newcommand{\tightlist}[1]{
\begin{itemize}[leftmargin=*, nosep, after=\strut, topsep=1.5pt]
#1
\end{itemize}\vspace{-\baselineskip}\mbox{}
}

\usepackage{colortbl}
\definecolor{Gray}{gray}{0.85}

\newcommand{\citepia}[1]{\ignorespaces\setcitestyle{notesep={; }}\citep[\emph{i.a.}]{#1}\ignorespaces\setcitestyle{notesep={, }}}

\newcommand{\highlight}[2]{%
\definecolor{hlcolor}{HTML}{#1} %
\sethlcolor{hlcolor}%
\hl{#2}%
}

\newcommand{\defer}[1]{%
  \highlight{FFD464}{#1}%
}

\newcommand{\defereq}[1]{%
  \mbox{\highlight{FFD464}{$#1$}}%
}

\newcommand{\M}{\mathcal{M}}

\newcommand{\ourmethod}{{Co}-\textsc{Llm}\xspace}
\newcommand{\baselinerandom}{{Co}-\textsc{Llm}-Random\xspace}
\newcommand{\baselinegreedy}{{Co}-\textsc{Llm}-Greedy\xspace}

\newcommand{\datasetgsm}{GSM8k\xspace}
\newcommand{\datasetmath}{MATH\xspace}
\newcommand{\datasetbioasq}{BioASQ\xspace}
\newcommand{\datasetalpacaeval}{AlpacaEval\xspace}
\newcommand{\datasetmtbench}{MT-Bench\xspace}
\newcommand{\datasettulu}{T\"ulu-v2-mixture\xspace}

\newcommand{\llama}[1][\empty]{%
    \textsc{Llama}\ifx#1\empty\xspace\else-\textsc{#1}\xspace\fi%
}
\newcommand{\llm}[1]{\textsc{#1}}

\newcommand{\highlightcolor}[2]{%
  \sethlcolor{#1}%
  \hl{#2}%
}

\definecolor{violet8}{HTML}{6741D9}
\definecolor{orange7}{HTML}{E67700}

\DeclareRobustCommand{\styledtext}[3]{\textcolor{#1}{\highlightcolor{#2}{\texttt{#3}}}}

\newenvironment{ttblock}{\ttfamily}{}

\definecolor{oc-gray-0}{HTML}{F8F9FA}
\definecolor{oc-gray-1}{HTML}{F1F3F5}
\definecolor{oc-gray-2}{HTML}{E9ECEF}
\definecolor{oc-gray-3}{HTML}{DEE2E6}
\definecolor{oc-gray-4}{HTML}{CED4DA}
\definecolor{oc-gray-5}{HTML}{ADB5BD}
\definecolor{oc-gray-6}{HTML}{868E96}
\definecolor{oc-gray-7}{HTML}{495057}
\definecolor{oc-gray-8}{HTML}{343A40}
\definecolor{oc-gray-9}{HTML}{212529}

\definecolor{oc-red-0}{HTML}{FFF5F5}
\definecolor{oc-red-1}{HTML}{FFE3E3}
\definecolor{oc-red-2}{HTML}{FFC9C9}
\definecolor{oc-red-3}{HTML}{FFA8A8}
\definecolor{oc-red-4}{HTML}{FF8787}
\definecolor{oc-red-5}{HTML}{FF6B6B}
\definecolor{oc-red-6}{HTML}{FA5252}
\definecolor{oc-red-7}{HTML}{F03E3E}
\definecolor{oc-red-8}{HTML}{E03131}
\definecolor{oc-red-9}{HTML}{C92A2A}

\definecolor{oc-pink-0}{HTML}{FFF0F6}
\definecolor{oc-pink-1}{HTML}{FFDEEB}
\definecolor{oc-pink-2}{HTML}{FCC2D7}
\definecolor{oc-pink-3}{HTML}{FAA2C1}
\definecolor{oc-pink-4}{HTML}{F783AC}
\definecolor{oc-pink-5}{HTML}{F06595}
\definecolor{oc-pink-6}{HTML}{E64980}
\definecolor{oc-pink-7}{HTML}{D6336C}
\definecolor{oc-pink-8}{HTML}{C2255C}
\definecolor{oc-pink-9}{HTML}{A61E4D}

\definecolor{oc-grape-0}{HTML}{F8F0FC}
\definecolor{oc-grape-1}{HTML}{F3D9FA}
\definecolor{oc-grape-2}{HTML}{EEBEFA}
\definecolor{oc-grape-3}{HTML}{E599F7}
\definecolor{oc-grape-4}{HTML}{DA77F2}
\definecolor{oc-grape-5}{HTML}{CC5DE8}
\definecolor{oc-grape-6}{HTML}{BE4BDB}
\definecolor{oc-grape-7}{HTML}{AE3EC9}
\definecolor{oc-grape-8}{HTML}{9C36B5}
\definecolor{oc-grape-9}{HTML}{862E9C}

\definecolor{oc-violet-0}{HTML}{F3F0FF}
\definecolor{oc-violet-1}{HTML}{E5DBFF}
\definecolor{oc-violet-2}{HTML}{D0BFFF}
\definecolor{oc-violet-3}{HTML}{B197FC}
\definecolor{oc-violet-4}{HTML}{9775FA}
\definecolor{oc-violet-5}{HTML}{845EF7}
\definecolor{oc-violet-6}{HTML}{7950F2}
\definecolor{oc-violet-7}{HTML}{7048E8}
\definecolor{oc-violet-8}{HTML}{6741D9}
\definecolor{oc-violet-9}{HTML}{5F3DC4}

\definecolor{oc-indigo-0}{HTML}{EDF2FF}
\definecolor{oc-indigo-1}{HTML}{DBE4FF}
\definecolor{oc-indigo-2}{HTML}{BAC8FF}
\definecolor{oc-indigo-3}{HTML}{91A7FF}
\definecolor{oc-indigo-4}{HTML}{748FFC}
\definecolor{oc-indigo-5}{HTML}{5C7CFA}
\definecolor{oc-indigo-6}{HTML}{4C6EF5}
\definecolor{oc-indigo-7}{HTML}{4263EB}
\definecolor{oc-indigo-8}{HTML}{3B5BDB}
\definecolor{oc-indigo-9}{HTML}{364FC7}

\definecolor{oc-blue-0}{HTML}{E7F5FF}
\definecolor{oc-blue-1}{HTML}{D0EBFF}
\definecolor{oc-blue-2}{HTML}{A5D8FF}
\definecolor{oc-blue-3}{HTML}{74C0FC}
\definecolor{oc-blue-4}{HTML}{4DABF7}
\definecolor{oc-blue-5}{HTML}{339AF0}
\definecolor{oc-blue-6}{HTML}{228BE6}
\definecolor{oc-blue-7}{HTML}{1C7ED6}
\definecolor{oc-blue-8}{HTML}{1971C2}
\definecolor{oc-blue-9}{HTML}{1864AB}

\definecolor{oc-cyan-0}{HTML}{E3FAFC}
\definecolor{oc-cyan-1}{HTML}{C5F6FA}
\definecolor{oc-cyan-2}{HTML}{99E9F2}
\definecolor{oc-cyan-3}{HTML}{66D9E8}
\definecolor{oc-cyan-4}{HTML}{3BC9DB}
\definecolor{oc-cyan-5}{HTML}{22B8CF}
\definecolor{oc-cyan-6}{HTML}{15AABF}
\definecolor{oc-cyan-7}{HTML}{1098AD}
\definecolor{oc-cyan-8}{HTML}{0C8599}
\definecolor{oc-cyan-9}{HTML}{0B7285}

\definecolor{oc-teal-0}{HTML}{E6FCF5}
\definecolor{oc-teal-1}{HTML}{C3FAE8}
\definecolor{oc-teal-2}{HTML}{96F2D7}
\definecolor{oc-teal-3}{HTML}{63E6BE}
\definecolor{oc-teal-4}{HTML}{38D9A9}
\definecolor{oc-teal-5}{HTML}{20C997}
\definecolor{oc-teal-6}{HTML}{12B886}
\definecolor{oc-teal-7}{HTML}{0CA678}
\definecolor{oc-teal-8}{HTML}{099268}
\definecolor{oc-teal-9}{HTML}{087F5B}

\definecolor{oc-green-0}{HTML}{EBFBEE}
\definecolor{oc-green-1}{HTML}{D3F9D8}
\definecolor{oc-green-2}{HTML}{B2F2BB}
\definecolor{oc-green-3}{HTML}{8CE99A}
\definecolor{oc-green-4}{HTML}{69DB7C}
\definecolor{oc-green-5}{HTML}{51CF66}
\definecolor{oc-green-6}{HTML}{40C057}
\definecolor{oc-green-7}{HTML}{37B24D}
\definecolor{oc-green-8}{HTML}{2F9E44}
\definecolor{oc-green-9}{HTML}{2B8A3E}

\definecolor{oc-lime-0}{HTML}{F4FCE3}
\definecolor{oc-lime-1}{HTML}{E9FAC8}
\definecolor{oc-lime-2}{HTML}{D8F5A2}
\definecolor{oc-lime-3}{HTML}{C0EB75}
\definecolor{oc-lime-4}{HTML}{A9E34B}
\definecolor{oc-lime-5}{HTML}{94D82D}
\definecolor{oc-lime-6}{HTML}{82C91E}
\definecolor{oc-lime-7}{HTML}{74B816}
\definecolor{oc-lime-8}{HTML}{66A80F}
\definecolor{oc-lime-9}{HTML}{5C940D}

\definecolor{oc-yellow-0}{HTML}{FFF9DB}
\definecolor{oc-yellow-1}{HTML}{FFF3BF}
\definecolor{oc-yellow-2}{HTML}{FFEC99}
\definecolor{oc-yellow-3}{HTML}{FFE066}
\definecolor{oc-yellow-4}{HTML}{FFD43B}
\definecolor{oc-yellow-5}{HTML}{FCC419}
\definecolor{oc-yellow-6}{HTML}{FAB005}
\definecolor{oc-yellow-7}{HTML}{F59F00}
\definecolor{oc-yellow-8}{HTML}{F08C00}
\definecolor{oc-yellow-9}{HTML}{E67700}

\definecolor{oc-orange-0}{HTML}{FFF4E6}
\definecolor{oc-orange-1}{HTML}{FFE8CC}
\definecolor{oc-orange-2}{HTML}{FFD8A8}
\definecolor{oc-orange-3}{HTML}{FFC078}
\definecolor{oc-orange-4}{HTML}{FFA94D}
\definecolor{oc-orange-5}{HTML}{FF922B}
\definecolor{oc-orange-6}{HTML}{FD7E14}
\definecolor{oc-orange-7}{HTML}{F76707}
\definecolor{oc-orange-8}{HTML}{E8590C}
\definecolor{oc-orange-9}{HTML}{D9480F}

\newcommand{\promptData}[1]{\styledtext{oc-violet-8}{oc-indigo-0}{#1}}

\usepackage[final, commandnameprefix=ifneeded]{changes}

\newcommand{\mitboston}{}%

\newif\iftrainingdetails
\trainingdetailsfalse

\title{Learning to Decode Collaboratively with Multiple Language Models}

\author{%
Shannon Zejiang Shen$^{\mitboston}$%
~\;~\;~Hunter Lang$^{\mitboston}$
~\;~\;~Bailin Wang$^{\mitboston}$ 
~\;~\;~Yoon Kim$^{\mitboston}$
~\;~\;~David Sontag$^{\mitboston}$\vspace{0.5ex}\\
    $^{\mitboston}$Massachusetts Institute of Technology \\
    \texttt{\href{mailto:zjshen@mit.edu}{\{zjshen, hjl, bailinw, yoonkim, dsontag\}@mit.edu}}
 } 

\begin{document}
\maketitle

\begin{abstract}

We propose a method to teach multiple large language models (LLM) to collaborate by interleaving their generations at the token level. 
We model the decision of which LLM generates the next token as a latent variable. By optimizing the marginal likelihood of a training set under our latent variable model, the \textit{base LLM} automatically learns when to generate itself and when to call on one of the ``assistant'' language models to generate, {all without direct supervision}.
Token-level collaboration during decoding allows for a fusion of each model's expertise in a manner tailored to the specific task at hand.
Our collaborative decoding is especially useful in cross-domain settings where a generalist base LLM learns to invoke domain expert models. %
On instruction-following, domain-specific QA, and reasoning tasks, we show that the performance of the joint system exceeds that of the individual models.
Through qualitative analysis of the learned latent decisions, we show models trained with our method exhibit several interesting collaboration patterns, e.g., template-filling.\footnote{Code:  \url{https://github.com/clinicalml/co-llm}}

\end{abstract}

\begin{figure}[!h]
  \centering
  \includegraphics[width=1\linewidth]{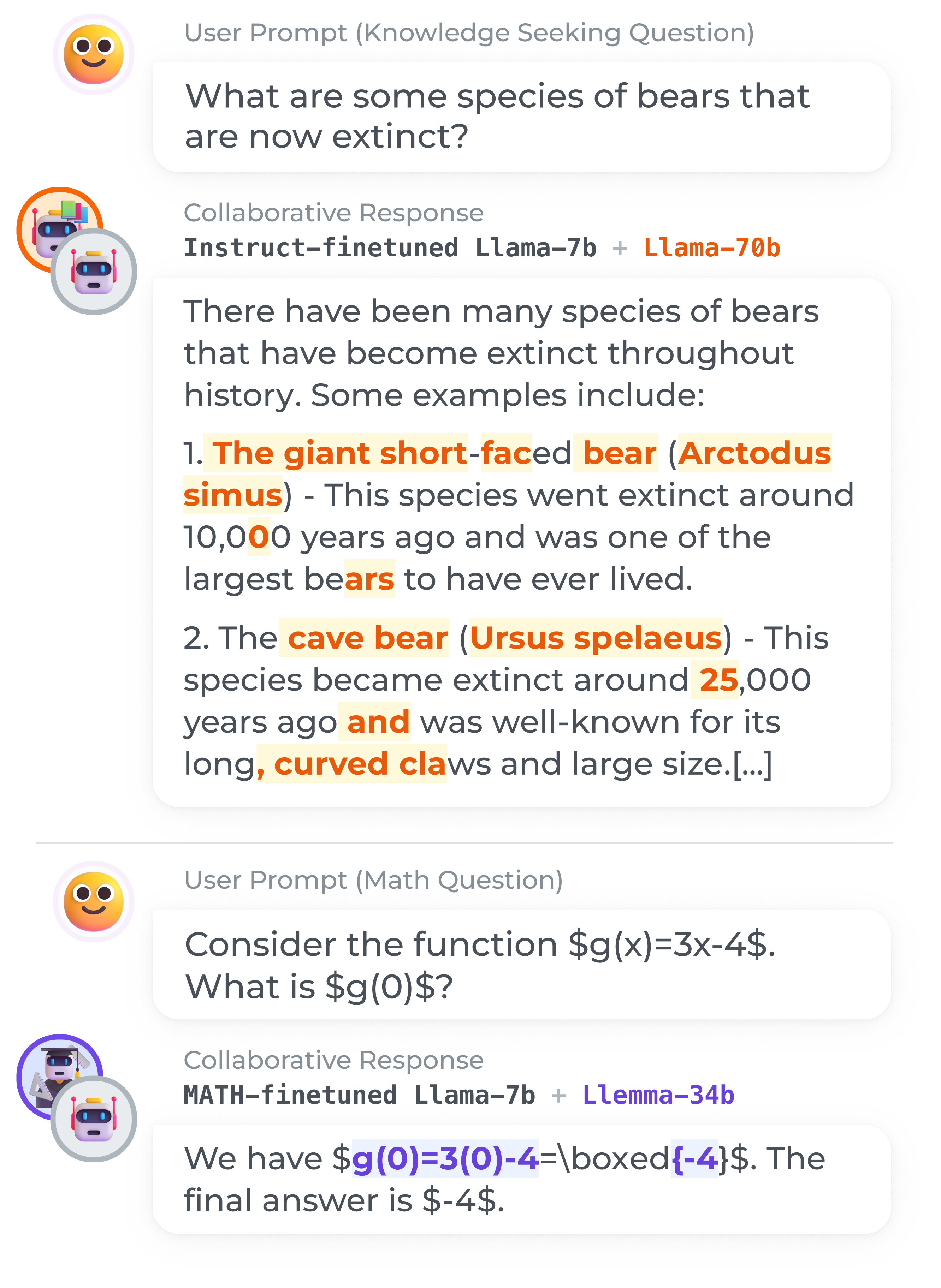}
  \vspace{-9.5mm}
  \caption{Example generations of our method, \ourmethod. 
 \textit{Top}: the base model generates the answer template and uses a larger \llm{Llama} model to fill in factual knowledge; \textit{Bottom}: the base model uses a math-specialized model as an ``API'' for computation.
The assistant model generated the highlighted tokens because the base model learned to \textit{defer} generation at those locations. 
  }
  \label{fig:instruction-following-example}
  \vspace{-6.5mm}
\end{figure}

\section{Introduction}

Techniques that combine the generations of multiple large language models (LLMs) at decoding time have benefits ranging from faster decoding speed \citep{leviathan2023fast}, to more controllable generations \citep{liu-etal-2021-dexperts, yang-klein-2021-fudge}, to more coherent, less repetitive text \citep{li-etal-2023-contrastive}, and even enabling a large model to be ``tuned'' by combining its generations with those of a smaller model from the same family \citep{liu2024tuning}.
A parallel thread of work has aimed to equip language models with the ability to infuse external tools into their generations, with the goal of incorporating outside knowledge and capabilities \citep{mialon2023augmented}.
Language models are able to produce more faithful and accurate generations when equipped with external APIs~\citepia{schick2023toolformer, qin2023toolllm}, search engines or retrievers~\citepia{izacard2022few, asai2023self, nakano2021webgpt}, or code executors~\citepia{gao2023pal}. 

\begin{figure*}[!t]
  \centering
  \includegraphics[width=1\linewidth]{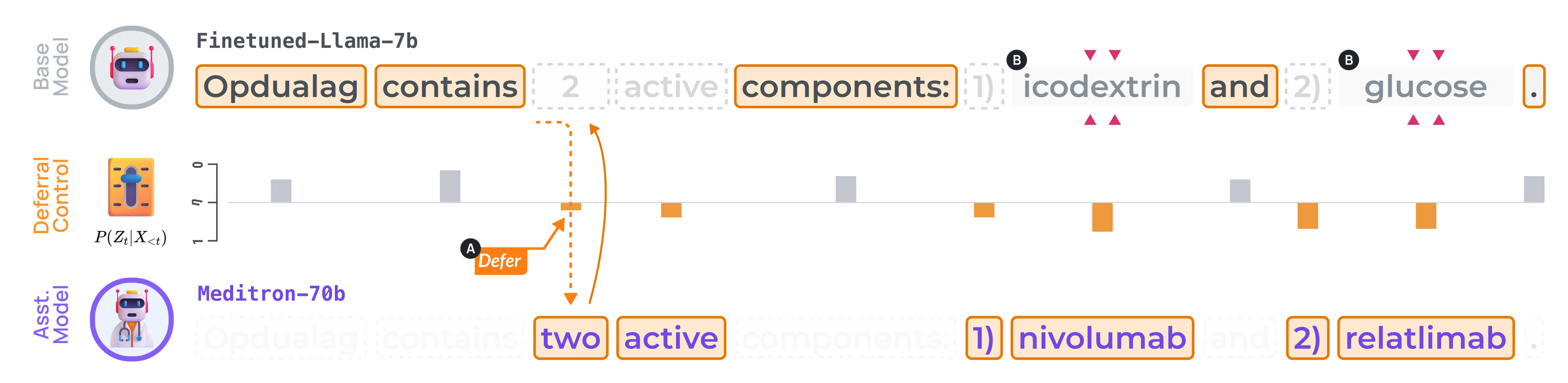}
  \vspace{-7.5mm}
  \caption{
  Illustration of the decoding procedure in \ourmethod, where a \styledtext{oc-gray-7}{oc-gray-1}{base} (\llama[7b]) and \styledtext{oc-violet-7}{oc-indigo-0}{assistant} model (\llm{Meditron}-70B) collaborate to generate a correct response for a medical question. 
  \added{For each token,} the \styledtext{oc-orange-7}{oc-yellow-0}{deferral control} \added{predicts the probability of switching to the assistant model to decode the next token given the context: it \emph{defers} when the probability is above some threshold $\eta$ (indicated by A), and uses the decoded token as the context (highlighted with orange border).} \deleted{predicts to invoke the assistant model given the current context when suitable, and interleaves the generations from both models: Tokens highlighted with an orange border constitute the final generation.}
  When using the base model alone, it may make factual mistakes (indicated by B); \ourmethod learns to use the assistant model at these positions to produce correct generations.
  }
  \vspace{-5.5mm}
  \label{fig:decoding-example}
\end{figure*}

While powerful, these methods all require a prescription on \textit{how} to combine the models and \textit{when} to use the tools, either via specific formulas for combining the logits of multiple models, or through (weak) supervision on where to insert tool/API calls in the training data.
In this work, we explore a different type of model combination where the models learn to {interleave} their generations token-by-token.
Each token is generated by one model, so the models \textit{collaborate} to generate a token sequence together.
We represent the decision of which LLM generates the next token as a latent variable, assuming {no direct supervision} on the decision of which model to use at each decoding step.
This enables an effective collaboration pattern for a given task to be learned organically from data.

Figure \ref{fig:instruction-following-example} shows example generations from our method, \ourmethod.
In the top example, \llama[7B] collaborates with \llama[70B] on instruction-following by generating a list template for the answer and then calling on \llama[70b] to fill in each element of the list.
Using the larger model as an assistant allows the smaller model to make effective use of a larger knowledge base and focus its efforts on learning the correct ``scaffolding'' for instruction responses.
In the bottom example, \llama[7B] collaborates with \llm{Llemma}-34B \citep[a domain-specific math model,][]{azerbayev2023llemma} by treating the latter as an API call to fill in parts of a LaTeX formula.
In both cases, the model predicts when to call the assistant by itself, behavior it \textit{learns} from training without direct supervision on which contexts suit the assistant model well.
This enables the emergence of qualitatively different collaboration methods (e.g., learning to scaffold, calling the large model as an API) based on what the task demands.

Section~\ref{sec:setup} describes our latent-variable model for collaboration during decoding, and Section~\ref{sec:method} describes the training and decoding procedures for \ourmethod in this model.
In, Section~\ref{sec:experiments}, we evaluate \ourmethod on instruction-following, mathematical reasoning, and domain-specific question-answering tasks.
Our results indicate that teaching models to collaborate improves performance across all of these tasks compared to using the individual models alone, and can sometimes match or exceed the performance of fine-tuning the large model.
By using chain-of-thought reasoning~\citep{wei2022chain}, \ourmethod can also be applied to classification tasks, where our experiments show that it boosts performance by enabling improved reasoning capability.
\deleted{Our results show that \ourmethod is especially useful in cross-domain settings where a generalist base LLM learns to invoke domain expert models.}
\added{Our results show that \ourmethod is especially useful in cross-domain settings where a generalist base LLM learns to invoke domain expert models and that \ourmethod can be effectively combined with other ensemble models, such as Mixture of Experts models \cite{shazeer2017outrageously}.}

\section{Latent-Variable Framework for Collaborative Generation}
\label{sec:setup}

Given a set of LMs with different expertise or sizes, we propose a latent-variable framework that enables their collaboration in a cost-efficient way.
The framework centers around a finetunable \textit{base model}, which itself is a relatively small LM. 
It decides which other \textit{assistant models} (which are typically larger and/or more specialized models) to use per token. 
When the base model calls on an assistant to generate the next token, we say it \textit{defers generation} for that token.

To generate a sequence of tokens $(X_1,\ldots,X_T)$, we represent the choice of which model generates token $X_t$ as a discrete latent variable $Z_t\in\{0, 1,\ldots$ $,M\}$, where $i=0$ denotes the base model, and $i \in \{1,\ldots,M\}$ refers to the $i$-th of $M$ assistant models.
We assume access to the conditional distributions $P_i(X_t | X_{<t})$, $i\in \{1,\ldots, M\}$, of the assistant models,\footnote{Conditionals can be obtained via either locally deployed models or remote API calls. At training time, we only require access to the assistant model's probability for the ground-truth tokens, not the full conditional distribution.} and full access to the base model. 
Using these distributions, we represent the \textit{joint sequence-level likelihood} as:
\begingroup\makeatletter\def\f@size{10}\check@mathfonts
\begin{align}
    & P(X, Z) = \prod_{t=1}^T \Bigl( P_{\theta}(Z_t|X_{<t}) P_{_{Z_t}}(X_t | X_{<t})\Bigr) 
    \label{eq:joint-likelihood} 
\end{align}\endgroup
The (learned) categorical distribution $P_\theta$ models the token-level discrete decision $Z_t$. %
For efficiency, each $Z_t$ is conditionally independent of $Z_{<t}$ given $X_{<t}$ in our design. %
The latent variable $Z_t$ is designed in the same spirit as the defer variable in \citet{mozannar2020consistent} and classical Mixture-of-Experts models \citep{jordan1994hierarchical} or ensemble models~\citep{Saunders2019DomainAI}. \deleted{or as in MoEs~\citep{jordan1994hierarchical,zhou2022mixtureofexperts,li2022branchtrainmerge}.}

\paragraph{Unsupervised Learning.} In practice, the token level decisions $Z_t$ are unknown, and collecting human annotation is not scalable. 
Our latent-variable framework offers a natural way of handling the issue with unsupervised learning. 
In particular, we aim to optimize the following {marginal likelihood},
\begingroup\makeatletter\def\f@size{9.5}\check@mathfonts
\begin{align}
    P(X) %
    &= \prod_{t=1}^T \Bigl(\sum_{Z_t=0}^M P_{\theta}(Z_t| X_{<t}) P_{_{Z_{t}}}(X_t | X_{<t})\Bigr),
    \label{eq:marg-likelihood}
\end{align}\endgroup
which can be computed efficiently during training due to the conditional independence structure. %

\paragraph{Collaborative Decoding.}
During inference time, our goal is to find the best sequence $X$ along with the best decision $Z$ on which assistant LM to use.
\begingroup\makeatletter\def\f@size{11}\check@mathfonts
\begin{align}
  \hat X, \hat Z &= \argmax_{X, Z}  P(X, Z)
  \label{eq:exact_decode}
\end{align}\endgroup
The exact $\argmax$ in \cref{eq:exact_decode} is intractable, so we follow the common practice of using the greedy strategy for decoding both $Z_t$ and $X_t$ in a token-by-token, autoregressive manner (see \cref{fig:decoding-example} for an example). In greedy decoding, for each token position, we first choose $\hat Z_t = \argmax_{Z_t} P_{\theta}(Z_t | X_{<t})$ to determine which model to decode from, then decode greedily from that model: $\hat X_t = \argmax_{X_t} P_{_{\hat{Z}_t}}(X_t | X_{<t})$.
Compared with standard greedy decoding for a single LM, decoding in this model is performed in collaboration, coordinated by $P_{\theta}$.
An alternative strategy to $\argmax$ decoding is to marginalize out $Z_t$:
\begingroup\makeatletter\def\f@size{11}\check@mathfonts
\begin{align}
  \hat X_t & = \argmax_{X_t} \sum_{Z_t} P_{_{Z_t}}(X_t | X_{<t})P_{\theta}(Z_t| X_{<t})
    \nonumber %
\end{align}\endgroup
which closely aligns with the marginal likelihood training objective.
However, this requires calling all $M+1$ models every token, slowing down decoding.
Our empirical results show that greedily choosing $Z_t$ based on $P_{\theta}(Z_t | X_{<t})$ performs well and enables interpretable collaboration, since each token is generated by a single model.

\paragraph{Remark.} The design of the collaborative decoding is natural from the probabilistic modeling perspective, and in this work we mainly focus on important empirical questions: how should we  parameterize $P_{\theta}$ so that it can be learned in a data-efficient way; can the base model learn how to cooperate with (larger or domain-specific) assistant models with only access to conditional probabilities (no internal hidden states or weights); and what kind of collaboration pattern can be induced if the collaboration is learned without supervision?
The latent-variable framework allows us to answer these questions via the exposed interpretable variable $Z_t$.

\section{\ourmethod: Learning to Decode Collaboratively with LMs}
\label{sec:method}

In this work, we focus on the basic case where we only have one base model and one assistant model. 
In this setting, the base model needs to make a binary decision of whether to generate from itself or defer to the assistant model, i.e., $Z_t \in \{0, 1 \}$. 
For clarity, we use $P_{\text{base}}$ and $P_{\text{asst}}$ to denote the base and assistant model, respectively. 
In the rest of this section, we explain our design for the parameterization of the model selector, and the training and inference procedure of the joint model.

\subsection{Modeling $P_\theta(Z_t | X_{<t})$} 

Since we have full access to the base model, we build $P_\theta$ on top of the base model for parameter efficiency.
Specifically, we represent $\theta$ as a linear binary classification head in the last layer. 
Formally, if $h_t(X_{<t}) \in \mathbb{R}^d$ is the base model's last hidden state at time step $t$ for inputs $X_{<t}$, and $\theta \in \mathbb{R}^d$ is the weight vector, we set:
\[
P_{\theta}(Z_t | X_{<t}) = \sigma(\langle \theta, h_t(X_{<t} \rangle),
\]
where $\sigma(\cdot)$ is the sigmoid function.
This introduces only $d$ new parameters to the base model, where $d$ is the base model's hidden dimension size. %

\subsection{Training}
\label{subsec:training}

Our training objective %
minimizes the negative log-marginal likelihood {\small $-\sum\nolimits_{t=1}^T \log P(X_t | X_{<t})$}, where 
\begingroup\makeatletter\def\f@size{10}\check@mathfonts
\begin{align}
    P(X_t | X_{<t}) = &P_{\text{base}}(X_t | X_{<t})P_{\theta}(Z_t = 0 | X_{<t}) + \notag \\ 
    &P_{\text{asst}}(X_t | X_{<t})P_{\theta}(Z_t = 1 | X_{<t}) \label{eq:joint-loss}
\end{align}\endgroup
is the likelihood of the next token after marginalizing out the latent variable $Z_t$.
The training procedure updates both $\theta$ and the base model parameters; it requires the forward pass of both the base and assistant model to obtain next-token probabilities, and the backward pass of the base model for gradient computation. 
This implies that any commercial LLMs or locally deployed LMs exposing next-probability output can be used as the assistant LMs. The marginal likelihood aligns well with the typical pretrained objective of maximizing the next-token probs. Moreover, offloading ``difficult tokens'' can potentially alleviate hallucination issues of the base model, thus leading to better generalization even without much help from the assistant model, as we will show in the experiments.

\paragraph{Initialization of $\theta$.} 
In our experiments, we found that appropriately initializing $\theta$ helps encourage the base LM to quickly switch from generating by itself to generating collaboratively.
Instead of collecting direct supervision for $Z_t$ values to initialize $\theta$, we use weak supervision to collect initial $Z_t$ values, use these pseudolabels to initialize the parameters $\theta$, then allow the $Z_t$'s to change during the training procedure.

Intuitively, the prime location for $Z_t = 1$ is when the assistant model correctly predicts the target token $X_t$ but the base model does not.
To that end, we set the pseudolabels for $Z_t$ as:
\begingroup\makeatletter\def\f@size{11}\check@mathfonts
\begin{align}
    \hat{Z}_t := \ind[&X_t=\argmax\nolimits_{v\in\mathcal{V}}P_{\text{asst}}(v | X_{<t}) \land \notag \\ 
    & X_t \ne \argmax\nolimits_{v\in\mathcal{V}}P_{\text{base}}(v | X_{<t})], \label{eq:initialization}
\end{align} \endgroup
and initialize the $d$ parameters $\theta$ by maximizing the likelihood of $\log P_\theta(\hat Z_t|X_{<t})$ while holding the rest of the base model fixed.
In our experiments, we compare to a baseline that combines this initialization with the usual language-model fine-tuning loss; our results show that the marginal likelihood objective leads to better performance by enabling the base model to learn better $Z_t$ values from data \added{(see ablation results in \cref{sec:different-scale-collaboration})}.

\subsection{Decoding}
\label{sec:model-decode}

In initial experiments, we found that performance of the joint model is sensitive to the choice of $Z_t$; due to the exposure of $Z_t$ in our latent-variable framework, we can impose extra priors over $Z_t$ for better performance. 
Specifically, we follow the general greedy decoding strategy, and set a threshold $\eta$ for decoding $Z_t$:
\begingroup\makeatletter\def\f@size{11}\check@mathfonts
\begin{align}
  \hat Z_t & =  \ind[P_{\theta}(Z_t = 1 | X_{<t}) > \eta],
    \label{eq:collab-decode-threshold}
\end{align}\endgroup
which means that when  $P_\theta(Z_t = 1) > \eta$, we execute the assistant model to predict the next token. The hyperparameter $\eta$ is picked via grid search on a small validation set per dataset. 
The choice of threshold for the decoding probability also allows for inference-time control over the amount of collaboration, in contrast with other approaches such as DExperts \citep{liu-etal-2021-dexperts}, and our performance degrades gracefully as the threshold increases.

\section{Experimental Setup}
\label{sec:experiments}

In our experiments, we fine-tune models for specific tasks  
and we test the models in-domain, comparing the end-task performance between \ourmethod and multiple single- or multi-model baselines. 
We test on 4 datasets ranging from instruction-following to solving expert problems, trying to understand when and how model collaboration can be beneficial.
We investigate the collaboration between different models (e.g., between Llama models of multiple scales, and between models fine-tuned on different domains). 
Overall, we find that {\ourmethod can learn a successful collaboration between different base and reference models, leading to better results than tuning base models alone.}

\paragraph{Models Used.} 
Our primary experiments are concerned with whether smaller models can collaborate with expert models that have been specialized to different domains.
In \cref{sec:domain-expert-collaboration}, we experiment with collaboration between the finetuned \llama[7B] and the \llm{Llemma} family~\citep[finetuned for math and reasoning]{azerbayev2023llemma}, as well as the \llm{Meditron} family~\citep[finetuned for biomedicine]{chen2023meditron70b}. 
It is also possible to use differently-sized models from the same family with \ourmethod: in \cref{sec:different-scale-collaboration} we experiment with the 7B and the 70B model from the same \llama[2] family~\citep{Touvron2023Llama2O} as the base and assistant model, respectively. 

\paragraph{Datasets.} We train on the full T\"ulu v2 mix data~\citep{Wang2023HowFC} for instruction following, \datasetgsm~\citep{cobbe2021training} and \datasetmath~\citep{hendrycks2021measuring}, each with 7.5k samples, for reasoning and math problem solving, and the \datasetbioasq~\citep{tsatsaronis2015overview} (4.7k samples) for medical question answering. 
We train and test the model on the corresponding data and evaluation suites separately. 
 
\paragraph{Evaluation.} 
We only compare greedy decoding results for all models, as they are commonly used in real-world applications.
We evaluate the instruction following model using the \datasetalpacaeval~\citep{alpaca_eval} benchmark; we use the GPT-4 annotator and compute the win rate of the testing model as judged by \texttt{GPT-4-0613} when compared to the outputs produced by \texttt{Davinci-003}. 
For \datasetgsm, \added{following~\citet{Wang2023HowFC},we extract the last numerical answer from the model's output, and calculate the exact match of the model prediction on a 200-example subset of the test set}.
The \datasetmath dataset provides 5 levels of math problems from 7 categories: similar to~\citet{wu2023empirical}, we skip the geometry category since it largely relies on Asymptote code, and sample 25 questions per level per category for evaluation, resulting in a 750-example subset. 
We adopt the prompting and evaluation code from \citet{azerbayev2023llemma} and \citet{lewkowycz2022solving}, extracting the last generated math after ``The answer is''. \footnote{\added{We use greedy decoding for \datasetgsm and \datasetmath; some literature refers to this as ``maj@1''.}}
The \datasetbioasq comes with 330 test examples of 4 categories: factoid, list, summary, and yes/no questions, evaluated using strict accuracy (SAcc.), F1, accuracy (Acc.), and Rouge-2 (R2). 
We test on 310 examples (saving 20 for validation) and reimplement the evaluation code.\footnote{See details in Appendix~\ref{apdx:evaluation-details}.}

\begin{table*}[!htb]

    \renewcommand{\arraystretch}{1.15}
    \setlength{\tabcolsep}{2.65pt}
    \begin{subtable}{.4\linewidth}
        \small 
        \resizebox{.98\linewidth}{!}{
        \begin{tabular}{lcc}
        \toprule
        
        \multicolumn{1}{r}{\textit{Math and reasoning tasks}}
        &
        \begin{tabular}[c]{@{}c@{}}\textbf{GSM}\end{tabular} &
        \begin{tabular}[c]{@{}c@{}}\textbf{MATH}\end{tabular} \\
        \midrule
        \llm{Llemma}-7B            & 4.0  & 2.0 \\
        \llm{Llemma}-34B           &  14.5  & 6.3 \\
        Finetuned \llama[7B]          & 34.5 & 7.6  \\
        Finetuned \llama[70B] {\footnotesize \ (QLoRA)}          & 52.5 & 11.7  \\
        \midrule
        PT (\llm{Llemma}-34B + \llm{Llama}-7B) & 30.0 & 20.9  \\
        PT (\llm{Llemma}-34B + \llm{Llemma}-7B) & 58.5 & 23.7  \\
        \midrule 
        \ourmethod-7B + \llm{Llemma}-7B    & 40.0 & 17.2  \\
        \ourmethod-7B + \llm{Llemma}-34B   & 43.5 & 24.5  \\
        \bottomrule
        \end{tabular}
        }
        \label{table:table1}
    \end{subtable}%
    \begin{subtable}{.6\linewidth}
        \small
        \raggedleft
        \resizebox{.98\linewidth}{!}{
        \begin{tabular}{lccccc}
        \toprule
        \multicolumn{1}{r}{\textit{\datasetbioasq tasks}}
        &
        \begin{tabular}[c]{@{}c@{}}\textbf{Factoid}\end{tabular} &
        \begin{tabular}[c]{@{}c@{}}\textbf{List}\end{tabular} &
        \begin{tabular}[c]{@{}c@{}}\textbf{Yes/No}\end{tabular} &
        \begin{tabular}[c]{@{}c@{}}\textbf{Summ.}\end{tabular} &
        \textbf{Avg.} \\
        \midrule
        \llm{Meditron}-7B            & 0.00  & 2.7 & 70.4 & 18.6 & 22.9 \\
        \llm{Meditron}-70B           & 17.2  & 16.1 & 80.2 & 21.1 & 33.7 \\
        Finetuned \llama[7B]               & 23.7 & 13.8 & 76.5 & 18.1 & 33.0 \\
        Finetuned \llama[70B] {\footnotesize \ (QLoRA)}          & 24.7 & 20.7 & 75.3 & 21.1 & 35.5  \\
        \midrule
        PT (\llm{Meditron}-70B + \llm{Llama}-7B) & 26.9 & 10.7 & 80.2 & 7.3 & 31.3 \\
        PT (\llm{Meditron}-70B + \llm{Meditron}-7B) & 26.9 & 23.5 & 82.7 & 11.0 & 35.6 \\
        \midrule 
        \ourmethod-7B + \llm{Meditron}-7B    & 17.2 & 16.0 & 72.8 & 19.8 & 31.4 \\
        \ourmethod-7B + \llm{Meditron}-70B   & 21.5  & 18.6 & 81.5 & 20.6 & 35.6 \\
        \bottomrule
        \end{tabular}
        }
        \label{table:table2}
    \end{subtable}
    \vspace{-1mm}
    \caption{\ourmethod enables collaboration between models trained on different domains: using the expert model trained for the domain (e.g., \llm{Llemma} for math and reasoning, and \llm{Meditron} for biomedical tasks) during decoding boosts performance compared to the fine-tuned base model, and sometimes performs even better than fine-tuned \llama[70b].
    Proxy Tuning~\citep[PT]{liu2024tuning} only performs well when all three of its component models ($\M, \M^+$, $\M^-$) are pretrained on the same domain mix.}
    \vspace{-4mm}
    \label{table:versatile_task_finetuning}
\end{table*}

\subsection{Baselines}
\label{sec:baselines}

\paragraph{Single models.} The performance of the base and assistant models can inform whether the learned collaboration is beneficial. 
We report 0-shot performance of the original untuned models and their finetuned counterparts. 
The same data and hyperparameters are used for model finetuning; for 70B models, we fine-tune using QLoRA~\cite{dettmers2023qlora} with the hyperparameters in Appendix~\ref{apdx:experiment-details}.

\paragraph{Other collaborative models.} 
We use two collaborative strategies from the literature.
Contrastive Decoding~\citep[CD]{Li2022ContrastiveDO, obrien2023contrastive} combines the output of the untuned ``expert'' (e.g., a 70B model) and ``amateur'' (e.g., a 7B model) models by subtracting their logits and sampling from the resulting distribution. 
We follow the setup in \citet{obrien2023contrastive}, setting the same $\alpha=0.1$ and $\beta=0.5$, and use unmodified \llama[70B] as the expert model and \llama[7B] as the amateur model.\footnote{In their paper, \citet{obrien2023contrastive} use a 1.5B parameter amateur model; as this model is not released, we use the 7B model instead. Different from their paper, we use 0- or 1-shot prompting in congruence with our other results.}
Proxy Tuning~\citep[PT]{liu2024tuning} proposes to approximate finetuning a (large) base model $\M$ by composing the outputs of smaller expert $\M^+$ and anti-expert models $\M^-$ with $\M$.
We include CD and PT results as ways to enhance untuned models and to simulate finetuning 70B models, respectively.
Both CD/PT require calling the smaller and larger models at each time step.
In contrast, \ourmethod may generate multiple tokens from the base model before calling the large model, which can be seen as a form of speculative decoding \citep{leviathan2023fast}.
For example, for a sequence of length $L$, Proxy Tuning makes $L$ calls\footnote{Here each ``call'' corresponds to a token-level decoding step, which is a sensible unit with which to measure inference latency as the context encoding portion can be easily parallelized and thus grows slowly with respect to context length in small batch regimes.} to the large model and $2L$ calls to the small model, whereas \ourmethod makes $fL$ calls to the large (assistant) model and $L$ calls to the small (base) model, where $0 \le f \le 1$ is the empirical frequency of deferral, the percent of $Z_t=1$ tokens.

\paragraph{Ablated \ourmethod.}
We consider different variants of \ourmethod to verify the necessity of a learned model selector $P_{\theta}$. First, we consider two simple heuristics as model selectors: \baselinerandom randomly chooses the base or the assistant model to produce a token with probability $p=0.5$; \baselinegreedy runs both models in parallel for each token and selects the token with the higher probability from either model.  This is a strong baseline since it requires observing next-token probabilities from both models at every decoding step.
Similar to our default setting, the base model is finetuned on target datasets while the assistant model is frozen. 

\paragraph{Weakly-supervised \ourmethod.} 
Finally, we consider a different weakly supervised training procedure for \ourmethod.  
This baseline is inspired by the process used to derive tool-use labels in Toolformer~\citep{schick2023toolformer}: the weak supervision for when to call the tool is chosen and \textit{fixed} before updating the language model parameters. 
Specifically, the training procedure is two-stage: first, we collect pseudo-labels $\hat{Z}_t$ using \cref{eq:initialization}.
Second, we jointly train $P_{\theta}(Z_t|X_{<t})$ and the base model by optimizing the weighted sum of $\log P(\hat{Z_t} | X_{<t})$ and the usual language modeling loss. 
This trains the base model to defer to the assistant when $\hat{Z}_t = 1$ while also fine-tuning it on the full training set. %
In contrast, our marginal likelihood training \cref{subsec:training} only uses the $\hat{Z}_t$ values to \textit{initialize} $\theta$ and allows the $Z_t$ values to evolve during training.  %

\section{Results}

\subsection{Collaboration across domains}
\label{sec:domain-expert-collaboration}

\cref{table:versatile_task_finetuning} shows that \ourmethod enables collaboration between \llama and domain-specific models and that this collaboration improves performance compared to the individual models themselves.
For example, being able to utilize \llm{Llemma} as an assistant leads to improved performances on math and reasoning tasks. 
On the \datasetmath dataset, even invoking a small, 7B-scale \llm{Llemma} assistant (17.2) outperforms fine-tuned \llm{Llama}-7B (7.6), fine-tuned \llm{Llama}-70B (11.7), and \llm{Llemma}-34B (6.3).
Similarly, cooperation with \llm{Meditron} models leads to performance gains on some BioASQ subtasks (e.g., List, Summary), and outperforms fine-tuned \llm{Llama}-7B, fine-tuned \llm{Llama}-70B, and base \llm{Meditron}-70B on average.
\begin{table*}[t]
    \resizebox{1.\linewidth}{!}{
    \begin{threeparttable}
        \renewcommand{\arraystretch}{1.12}
        \setlength{\tabcolsep}{3pt}
        \begin{tabular}{cllclclclccccc}
            \toprule
             & & &
             \textbf{AlpacaEval} &
              &
             \textbf{GSM} &
              &
             \textbf{MATH} &
              &
             \multicolumn{5}{c}{\textbf{BioASQ}\tnote{a}} \\
             \cmidrule{4-4} \cmidrule{6-6}  \cmidrule{8-8} \cmidrule{10-14}
                                                                 & &  &(\% Win) &  &(Acc.) &  &(EM) &  & {\small Factoid} (SAcc.) & {\small List} (F1) & {\small Yes/No} (Acc.) & {\small Summ.} (R2) & Avg.  \\
             \midrule
             \multirow{3}{*}[0ex]{\rotatebox{90}{\textit{Untuned}}} 
             & & \llama[7B]                                          & -    &  & 7.0  &  & 0.3    &  & 4.3           & 4.9      & 71.6         & 17.2      & 24.5 \\
             & & \llama[70B]                                         & 11.6  &  & 13.5 &  & 2.1  &  & 11.8          & 14.9     & 77.8         & 18.6      & 30.8 \\
             & & \llama[70B+7B] {\footnotesize \ (CD)}               & - &  & 11.5 &  & 1.3 &  & 11.8 & 9.0 & 71.6 & 17.5 & 27.5 \\
             \midrule
             \multirow{3}{*}[0ex]{\rotatebox{90}{\textit{Finetuned}}} 
             & & \llama[7B]{\footnotesize \ (Finetuned)}             & 69.3 &  & 34.5 &  & 7.6  &  & 23.7          & 13.8     & 76.5         & 18.1      & 33.0 \\
             & & \llama[70B]{\footnotesize \ (QLoRA)}                & 78.6\tnote{b} &  & 52.5 &  & 11.7 &  & 24.7          & 20.7     & 75.3         & 21.1      & 35.5 \\
            & & \llama[70B+7B]{\footnotesize \ (PT)}                     & 72.3 &  & 52.5 &  & 17.3 &  & 29.0 & 16.8 & 85.2 & 21.3 & 38.1 \\
             \midrule
             \multirow{5}{*}[-.5ex]{\rotatebox{90}{\textit{Collaboration}}} 
             & & \textsc{Co}-Random                                  & 46.3 &  & 17.0 &  & 6.1 &  & 6.5          & 1.9     & 30.9         & 17.5      & 14.3 \\
             & & \textsc{Co}-Greedy                                  & 64.1 &  & 38.0 &  & 8.1 &    & 29.0          & 16.6     & 76.5         & 20.2      & 35.6 \\
             & & Weak Supervision                                    & 56.7  &  & 40.0 &  & 12.3 & & 22.6          & 14.6      & 80.2        & 17.5      & 33.7  \\
             \cdashline{3-14}[.4pt/1pt]\noalign{\vskip 1pt}
             & & \ourmethod-7B (Base Only)                           & 70.6 &  & 33.0 &  & 6.4 &  & 20.4          & 11.2       & 79.0             & 18.1      & 32.2 \\
             & & \ourmethod-7B + \llama[70B]                         & 71.9 &  & 45.0 &  & 15.1 &  & 24.7          & 18.0     & 82.7         & 20.4      & 36.5 \\
             \bottomrule
        \end{tabular}
        \begin{tablenotes}
            \item[a] For BioASQ, we use 1-shot prompting for \llama[7B], -70B, and CD experiments to inform the model of the output format. 
            \item[b] We report the results obtained by \citet{ivison2023camels} in Table 5.
        \end{tablenotes}
    \end{threeparttable}
    }
\vspace{-2mm}
\caption[Caption for LOF]{\replaced{Results of using \ourmethod for \llama models of different sizes}{{\footnotesize Task-specific finetuning results}}. Occasionally\protect\footnotemark \ calling the Llama-70B model to generate a few tokens, \ourmethod-7B is able to significantly outperform the finetuned Llama-7B model in all tasks, and sometimes even performs better than the QLoRA-finetuned Llama-70B model.}
\vspace{-4mm}
\label{table:task_finetuning_results}
\end{table*}

\added{In addition, \ourmethod with \llm{Llama}-7B and \llm{Llemma}-34B can achieve similar performance as fine-tuned \llm{Llemma}-7B, which scores 43.5 on \datasetgsm and 23.5 on \datasetmath.
\ourmethod allows the base 7b model access to collaborate with a domain expert (\llm{Llemma}-34B), which surprisingly leads to similar performance as performing a large amount of domain-specific fine-tuning plus further task-specific fine-tuning on the base model (finetuned \llm{Llemma}-7b).}

These results suggest that \ourmethod enables a modular approach to continued pretraining and task-specific finetuning: one can pretrain a large model on a domain-specific corpus, then fine-tune smaller models with \ourmethod to leverage the knowledge from the larger models and attain improved performance on the downstream tasks. 

\begin{table}[!tb]

    \renewcommand{\arraystretch}{1.15}
    \setlength{\tabcolsep}{2.65pt}
    \centering
    \small 
    \begin{tabular}{lcc}
    \toprule
    
    \multicolumn{1}{r}{\textit{Math and reasoning tasks}}
    &
    \begin{tabular}[c]{@{}c@{}}\textbf{GSM}\end{tabular} &
    \begin{tabular}[c]{@{}c@{}}\textbf{MATH}\end{tabular} \\
    \midrule
    \llm{Mistral}-7B            & 21.5   & 7.2 \\
    \llm{Mixtral}-8$\times$7B (MoE) & 38.5   & 16.2 \\
    Finetuned \llm{Mistral}-7B & 51.0 & 13.9  \\
    \midrule 
    \ourmethod \llm{Mistral}-7B + \llm{Mixtral}-8$\times$7B & \textbf{57.0} & \textbf{20.0}  \\
    \bottomrule
    \end{tabular}
    \caption{\ourmethod can be applied among models of different architectures like a dense LLM (\llm{Mistral}-7B) and a sparse Mixture of Experts (MoE) model (\llm{Mixtral}-8$\times$7B). The learned collaboration leads to strong performance improvements on both GSM and MATH tasks.}
    \label{table:mistral-results}
    \vspace{-5mm}
\end{table}

\paragraph{Comparison against Proxy Tuning.}
While PT and our work are differently motivated and constructed, they both leverage multiple models during generation. \cref{table:versatile_task_finetuning} also provides an in-depth comparison between the two methods in the context of combining models from different domains. 
PT only performs well when all three models ($\M, \M^+$, $\M^-$) are pretrained on the same domain mix (compare, e.g. ``\llm{Llemma} + \llm{Llama}'' to ``\llm{Llemma + Llemma}'').
This is due to the implicit assumption that the difference between the base model $\M$ and a hypothetical, tuned version of the base model is the same as the difference between the smaller expert $\M^+$ and the anti-expert $\M^-$. 
Our results show that \ourmethod is more effective at enabling collaboration between models from different domains. 
PT also requires more calls to the larger model, thus resulting in slower inference. \ourmethod makes fewer calls to both large and small models.

In the following section, we show that in addition to enabling collaboration across domains, \ourmethod also allows collaboration across model scales.

\begin{table*}[t]
    \centering
    \resizebox{1.\linewidth}{!}{
    \small
    \renewcommand{\arraystretch}{1.0}
    \setlength{\tabcolsep}{4pt}
    \begin{tabular}{p{0.2\textwidth}p{0.2\textwidth}p{0.2\textwidth}p{0.2\textwidth}p{0.2\textwidth}}
    \toprule
    \multicolumn{1}{c}{Deferral frequency $f=0$}
    &
    \multicolumn{1}{c}{$f=0.3$}
    &
    \multicolumn{1}{c}{$f=0.4$}
    & 
    \multicolumn{1}{c}{$f=0.9$}
    &
    \multicolumn{1}{c}{$f=1.0$}
    \\
    \midrule
    We have $a^3 = 5^3 = 125$, and $a^2 = 5^2 = 25$, so $a^3\cdot a^2 = \boxed{125}$. The final answer is $125$.
    & 
    We have $a^{\defereq{3}} = \defereq{5^3} = \defereq{12}5$, and $a^2 = \defereq{5^2} = \defereq{2}5$, so $a^{\defereq{3}}\cdot a^2 = \boxed{\defereq{3}000}$. The final answer is $3000$.
    & 
    We have $a^{\defereq{3}}$ $=$ $\defereq{5^3}$ $=$ $ \defereq{12}5$, and $a^2$ = $\defereq{5^2} = \defereq{25}$, so $a^{\defereq{3}}\defereq{\cdot} a^2 =$ $\boxed{\defereq{312}5}$. The final answer is $3125$.
    & 
    We \defer{can use the power rule to simplify this expression.} We \defer{have that $a^3\cdot a^2 = a^{3+2} = a^5$}.  \defer{Now we can substitute $a=5$ to get $5^5 = 3125$.  Therefore, the final} answer is $\boxed{\defereq{3125}}$. The final answer is $3125$.
    & 
    \defer{\#\#\# 1 Given a mathematics problem, determine the answer. Simplify your answer as much as possible. You can use latex to format your answer and you should state your final answer as "The final answer is \$(final - answer)\$." Problem:}
    \\ 
    \bottomrule 
    \end{tabular}
    }
    \caption{Model answers with different rates of deferral to the question ``Evaluate the expression $a^3\cdot a^2$ if $a= 5$'' (answer: $3125$). In this example, we use the finetuned \llama[7B] as the base model and \llm{Llemma}-34B model as the assistant. We show the 0-shot model answers at different deferral frequencies $f$, with yellow background to indicate the token is produced by the assistant model.}
    \label{table:exampe-1}
    \vspace{-4mm}
\end{table*}  

\subsection{Collaboration across scales} 
\label{sec:different-scale-collaboration}
\cref{table:task_finetuning_results} shows that using \ourmethod leads to a successful collaboration between the base and assistant models of different sizes within the \llama family.
In particular, compared to using the (finetuned) base model \llama[7B] alone, \ourmethod-7B + \llama[70B], which occasionally calls the unmodified assistant model during decoding, consistently achieves significantly better performance across all tasks and datasets (2.6, 10.5, 7.5, and 3.3 absolute improvements for the 4 datasets, respectively).
\ourmethod is sometimes better than the QLoRA finetuned assistant model (on \datasetmath and \datasetbioasq), suggesting that our method can effectively combine the best of the models and achieve better performance than the ``sum'' of their parts. 
Training with \ourmethod does not hurt the performance of the base model: when we prohibit using the assistant model during inference, performance is comparable to the base model finetuned with the usual language modeling objective (\llama[7B] (Finetuned)): for example, getting 33.0 and 6.4 for \datasetgsm and \datasetmath, respectively.
\ourmethod thus degrades gracefully as the amount of deferral is changed, which we explore further in \S\ref{sec:deferral-frequency}.

Comparing with the model collaboration baselines, we show that interleaving two model generations is not a trivial task: randomly switching between the two models lead to worse than single-model performances in \baselinerandom.
Even if running two models in parallel, \baselinegreedy does not consistently yield better performance than using either model alone, and in some cases, they are worse (e.g., in the case of \datasetgsm).
As discussed in Section \ref{sec:domain-expert-collaboration}, Proxy Tuning performs very well when used with models of different sizes in the same family, but \ourmethod also performs well despite using far fewer calls to the language models.
Finally, the Toolformer-style baseline with fixed, weakly supervised values of $Z_t$ leads to overall worse performance compared to \ourmethod, indicating the benefits of our latent variable formulation and marginal likelihood loss, which allow the best deferral patterns to be learned from data.

\subsection{\added{Collaboration across architectures}}
\label{sec:different-architecutre-collaboration}

\added{Since \ourmethod only assumes access to token logprobs, it can easily be used for collaboration between models of different architectures. 
For example, \cref{table:mistral-results} shows that \ourmethod can be adopted to collaborate between a dense model (\llm{Mistral}-7B) and a sparse MoE model (\llm{Mixtral}-8$\times$7B), and the joint model achieves strong accuracy gains compared to either the finetuned \llm{Mistral}-7B model or the \llm{Mixtral}-8$\times$7B model.
These results indicate that \ourmethod is still useful when used together with other model combination methods like MoE.}

\begin{figure}[t]
    \centering
    \includegraphics[width=0.95\linewidth]{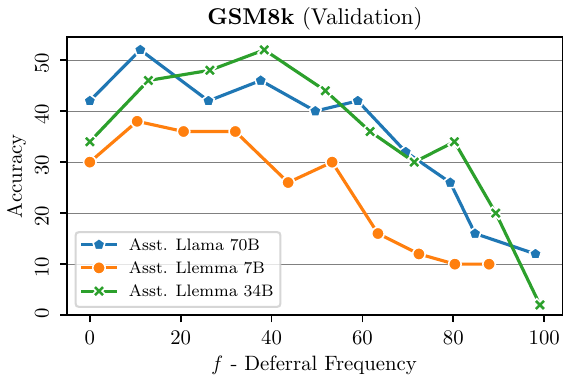}
    \label{fig:gsm8k}
    \caption{Performance of \ourmethod at different frequencies of deferral on \datasetgsm. There exists an optimal $f$ that the joint model achieves better performance than using either of them alone. Similar trend is observed in \datasetmath and \datasetbioasq, shown in \cref{fig:thresholds-acc-more} in Appendix.}
    \label{fig:thresholds-acc}
    \vspace{-2mm}
\end{figure}

\footnotetext{See \cref{sec:deferral-frequency} for a detailed analysis of the frequency of invoking the assistant model and the end performance.}

\subsection{Qualitative Analysis}
\label{sec:deferral-frequency}

The exposure of the interpretable variable $Z_t$ in our latent-variable framework makes it easy to visualize the model composition patterns.  As shown in \cref{fig:instruction-following-example,fig:decoding-example}, our unsupervised training can induce an interesting template-filling collaboration strategy: the base model learns to generate a template with slots that requires deeper knowledge (in question answering) or reasoning (in math problems) for assistant models to fill in.
\cref{table:exampe-1} further illustrates how the collaboration between the assistant (\llm{Llemma}-34B) and base (fine-tuned \llama[7B]) model leads to correct responses while either model alone fails to do so.
When the base model is tasked to solve the math question alone (i.e, deferral frequency $f=0$), it fails to produce a valid answer, generating $125$ rather than $3125$. 
As we increase the frequency of deferral by lowering the threshold $\eta$, we observe \ourmethod starts to invoke the assistant model to generate latex code and compute the results, and when $f=0.4$, the joint model produces the correct answer. 
More deferral in this case does not yield better generations. %
When we fully rely on the assistant model ($f=1$), since it is not tuned or aligned, it produces no helpful solutions.

We also evaluate the joint model at different deferral frequencies on small validation sets for \datasetgsm, \datasetmath, and \datasetbioasq,
and plot the results in \cref{fig:thresholds-acc} and \cref{fig:thresholds-acc-more} in the Appendix.
Across different domains and scales, the models exhibit a similar concave performance curve: at some optimal deferral frequencies, the joint model can achieve better performance than using either of the models alone, with decreased performance at both extremes.
The optima vary across different datasets, corresponding to the different patterns of invoking assistant models (e.g., API-call or ``leading'' style).
In practice, one can pick the proper $\eta$ %
to balance the accuracy and the efficiency/cost.

\section{Related Work}

\paragraph{Learning to compose models.}
Composing expert models has been a recurring strategy to improve large models, and the way of composing expert models is largely related to the underlying learning settings.  Mixture of Experts~\cite[MoE]{shazeer2017outrageously,jiang2024mixtral,  Dai2024DeepSeekMoETU,Xue2024OpenMoEAE} requires that all experts are trained simultaneously using the same data. 
In contrast, post-hoc methods assume the presence of pretrained language models but usually pose implicit constraints on the model families to compose. 
Proxy Tuning~\citep{liu2024tuning, liu-etal-2021-dexperts} works best when all models are all pretrained on the same data mixture; PHATGOOSE~\citep{muqeeth2024learning} requires that all models are LoRA-finetuned~\citep{hu2021lora} models; 
Contrastive Decoding~\citep[CD]{Li2022ContrastiveDO} requires an amateur model, which is not clear for tasks such as math reasoning. \ourmethod is more flexible, mainly because our base model \textit{learns} to interact with assistant models instead of relying on prescribed strategies.
Our experiments also indicate that Proxy Tuning (concurrent work) only performs well when all models are pretrained in the same domain, whereas \ourmethod can effectively combine general and domain-specific models.
Compared to CD and Proxy Tuning, \ourmethod also makes fewer calls to the language models at inference time, as described in Section \ref{sec:baselines}.

Perhaps most similar to our work are speculative decoding \citep{leviathan2023fast} and CombLM \citep{ormazabal2023comblm}.
Like our approach, speculative decoding generates some tokens with one model and some tokens with another, but our method differs in that the goal is to improve generation quality rather than to sample more quickly from a large model by approximating its generations with a smaller one.
CombLM also learns to combine the probability distributions of multiple models, but their combination is not time-varying, and they mainly demonstrate wins on perplexity.

Our approach could be seen as a special case of Toolformer~\citep{schick2023toolformer}, where the assistant model is the tool/API called by the base model. 
However, our latent variable formulation allows fine-grained deferral decisions %
to evolve over the course of fine-tuning, whereas Toolformer derives a fixed dataset prescribing which tool gets called (and where) \textit{before} training.
\ourmethod enables varying the frequency of calling the assistant model, whereas Toolformer has no provision for flexibly adjusting the amount of tool use at inference time.

\paragraph{Is this just Mixture of Experts?} \added{In Mixture of Experts (MoE) LLMs~\citep{zhou2022mixtureofexperts,li2022branchtrainmerge, jiang2024mixtral,Xue2024OpenMoEAE, Dai2024DeepSeekMoETU}, the goal is to train a model that can be partially executed during inference time.
A common choice in MoEs is to decompose FFN layers (feed forward network) in a transformer into modular ``experts''. 
These experts are subnetworks---they cannot be used standalone, and the experts are typically expected to have the same size (parameter count).
MoE also requires gradient access to all experts and assumes every expert has access to the same training data.}

\added{In contrast, we aim to learn how to collaborate with existing off-the-shelf models. First, we do not assume gradient access to assistant models, and we only fine-tune the base model. 
This saves in training cost, but poses a more difficult learning problem, since we only assume access to logprobs from the assistant model. 
\ourmethod is able to learn effective collaboration patterns between models despite this limited access to the assistants.}
\added{Second, because of our less restrictive assumptions on assistant model access and architecture, \ourmethod can be applied in more constrained scenarios. 
For example, \ourmethod can still be applied when the assistant is trained on proprietary data (e.g., due to copyright issues \citep{duetting2023mechanism}).}
\added{Third, our framing allows flexible collaboration between models of different sizes or even architectures. Table \ref{table:mistral-results} shows performance gains from combining a dense model with an MoE model using \ourmethod, indicating that the gains are orthogonal to MoE.}
\added{Finally, in our approach, each ``expert'' is a full-fledged LLM rather than a sub-network.
This allows for interpretable generation patterns (as illustrated in \cref{fig:instruction-following-example,fig:decoding-example}), as well as flexible inference-time tradeoffs between how much each model is used (e.g., in \cref{table:exampe-1}, we show \ourmethod can work well at different frequencies of deferral).
}

\paragraph{Learning to defer.}
A large body of literature focuses on the related problems of \textit{prediction with rejection} \citep{cortes2016learning}, where the goal is to train a model that can predict on some inputs and decline to predict on others, and \textit{learning with deferral} \citep{mozannar2020consistent, mozannar2023should}, where the goal is to train a model to make predictions on some inputs and defer to a human expert on others.
\citet{mohri2023learning} combine prediction with example-level rejection and LLMs on the text decontextualization problem.
We use a latent variable formulation inspired by \citet{mozannar2023should}, initially developed for learning to defer to a human expert on classification problems.
We replace the human expert with a fixed LLM assistant model and extend the loss to allow for token-level rather than sequence-level deferral.

\section{Conclusion}

We propose \ourmethod, a latent variable approach that learns to collaboratively %
decode using multiple LMs. 
\ourmethod can interleave token generations from different LMs in patterns most suitable for the task at hand (e.g., scaffolding the generation, or using the other model like an API), and learns the collaboration pattern organically from data.
We empirically show that \ourmethod can produce generations of better quality compared to using either of the models alone, in tasks ranging from math reasoning, medical question answering, and instruction following. 
The latent ``defer'' variable offers a flexible and interpretable way for adjusting the frequency for invoking other LMs at inference time without re-training. In the future, we plan to extend \ourmethod to integrate more than two LMs and investigate potentially more complex collaboration strategies emerging from this setting.

\section*{Limitations}
As shown in \cref{fig:thresholds-acc-more} and \cref{table:evaluation_setup} of the Appendix, the optimal deferral threshold may be different across different datasets and models. 
\ourmethod thus requires picking the deferral threshold per task, which can be inconvenient in practice.
However, the threshold also enables inference-time control over the amount of collaboration.
Second, not every deferral matters: for some positions, the assistant model may generate an identical token as the base model does. 
This suggests the development of more fine-grained control of deferral strategies, potentially via more sophisticated modeling of deferral model parameters $\theta$. 

Another limitation in our method comes from fully relying on an assistant model at some point in decoding. 
For example, if the assistant model is not well-tuned or aligned, it may unintentionally break the generation due to occasional mistakes. 
As shown in the example below, 
one erroneous token might lead to a cascade of errors, causing repetition patterns or generating irrelevant content. One future work is to develop a more robust deferral strategy that allows backtracking when the assistant model fails to generate a proper response.

\vspace{2mm}

{\small 
\begin{ttblock}
\noindent 
Here's a recipe for Kubdari, a traditional Georgian dish:

\noindent 
Ingredients:

\noindent 
* 1 lb ground beef

\noindent 
*\defer{1} onion, finely cho\defer{pped}

\noindent 
* 2 cloves\defer{garlic}, minced

\noindent 
*\defer{1} cup\defer{ch}opped\defer{parsley}

\noindent 
*\defer{1} cup chopped\defer{cilantro}

\noindent 
*\defer{1} cup chopped\defer{dill}

\noindent 
*\defer{1} cup chopped ...
    
\end{ttblock}
\noindent \emph{[...repeating the same pattern...]}
}

\section*{Acknowledgements}
SS and DS were supported by the National Science Foundation (NSF award no. IIS-2205320 Conceptualizing ML for Dynamic Information Retrieval of EHR Notes).
HL was funded by an NDSEG fellowship.
BW and YK were partially supported by MIT-IBM Watson AI Lab and an Amazon award. 
We thank CloudBank~\cite{norman2021cloudbank} for supplying computational resources, which is supported by the National Science Foundation under award \#1925001.
And thanks to Jacob Andreas, Lucas Torroba Hennigen, Hussein Mozannar, Ilker Demirel, Andreas Haupt, Tiwalayo Eisape, Alex Gu, Ruochen Zhang, and Doug Downey for feedback on the draft of this paper.

\bibliography{anthology,custom}

\newpage
~
\appendix

\section{Additional Experimental Details}
\label{apdx:experiment-details}
\label{apdx:evaluation-details}

\subsection{Model Training and Computation}

\paragraph{Training details.}
\iftrainingdetails
\else
In our experiments, we finetune the base model and learn the latent variable parameters $\theta$ jointly: i.e., we optimize both the ``$\text{base}$'' and $\theta$ parameters by optimizing the marginal likelihood from \cref{eq:joint-loss} with AdamW \citep{loshchilov2018decoupled}. 
We train our models primarily using 4 A100 80G GPUs, and we use FlashAttention~\citep{dao2023flashattention2} and DeepSpeed ZeRO Stage 2~\citep{rasley2020deepspeed} during training to reduce the GPU memory usage. 

We follow \citet{ivison2023camels} and \citet{liu2024tuning}, using similar hyperparameters and settings detailed in \cref{table:hyperparameters}.
In all training experiments, we compute the marginal likelihood loss on only target tokens (i.e., we mask out the input tokens). 
It takes around 2 hours to finish the finetuning experiments for \datasetgsm, and we estimate a total of 3,000 GPU hours used for all experiments.

\paragraph{Datasets and prompts.} \cref{table:dataset} details the sizes of the training and evaluation datasets. 
We format the data using the same simple prompts during training and evaluation, with examples shown in \cref{table:prompts}.

\fi

\paragraph{Model and data licenses.} We use three different LLMs in our experiments, i.e., \llm{Llama}~\cite{Touvron2023Llama2O}, \llm{Llemma}~\citep{azerbayev2023llemma}, and \llm{Meditron}~\citep{chen2023meditron70b}, and they all share the same LLaMA 2 community license.\footnote{\url{https://ai.meta.com/llama/license/}.}
The licenses for the dataset are listed in \cref{table:dataset-licenses}.

\begin{figure}[!h]
    \centering
    \begin{subfigure}[b]{0.95\linewidth}
      \includegraphics[width=\textwidth]{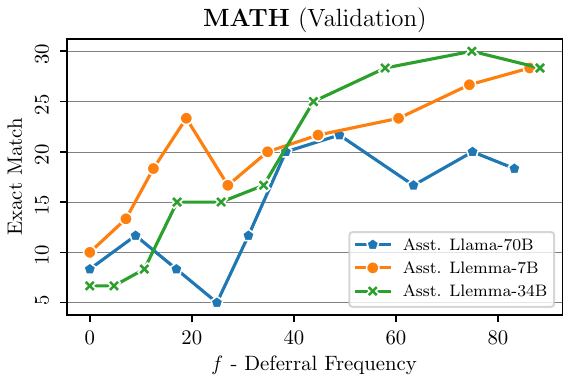}
      \label{fig:math}
    \end{subfigure}
    \vspace{-5mm}
    \begin{subfigure}[b]{0.95\linewidth}
        \includegraphics[width=\textwidth]{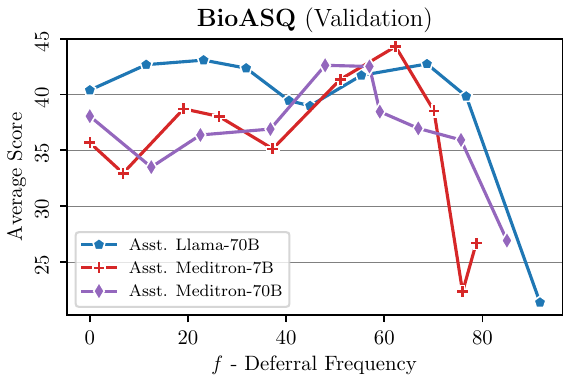}
        \label{fig:bioasq}
    \end{subfigure}
    \vspace{-2mm}
    \caption{Performance of \ourmethod at different frequencies of deferral on \datasetgsm, \datasetmath and \datasetbioasq. There exists an optimal $f$ that the joint model achieves better performance than using either of them alone.}
    \label{fig:thresholds-acc-more}
    \vspace{-2mm}
\end{figure}

\paragraph{Threshold search.}
For all methods with a learnable model selector (including \ourmethod), we choose the best $\eta$ for \cref{eq:collab-decode-threshold} using a small validation set before testing, described in \cref{alg:search-deferral-threshold}.
For \datasetgsm, \datasetmath, and \datasetbioasq, we conduct the search in-domain, using a small hold-out subset for that dataset; for \datasetalpacaeval, we use a subset of a separate instruction following benchmark \datasetmtbench~\citep{zheng2023judging} to search and pick the $\eta*$.
We report the used values under different settings in \cref{table:evaluation_setup} (for \ourmethod) and \cref{table:evaluation_setup} (for Weakly-supervised \ourmethod).

\begin{table}[!t]
    \resizebox{1.\linewidth}{!}{
    \centering
    \begin{threeparttable}
        \renewcommand{\arraystretch}{1.1}
        \setlength{\tabcolsep}{4pt}
        \begin{tabular}{lr}
        \toprule
        \textbf{Hyperparameter}      & \textbf{Configuration} \\
        \midrule
        \multicolumn{2}{c}{Default and \ourmethod Finetuning for 7B Models\tnote{a}} \\ 
        \midrule
        Training Epoch               & 2 \\ 
        Max Sequence Length          & 2048 \\
        Effective Batch Size         & 128 \\
        Gradient Accumulation Steps  & 16 \\
        \cdashline{1-2}[.4pt/1pt]\noalign{\vskip 1pt}
        Learning Rate                & 2e-5 \\
        Warmup Ratio                 & 0.04 \\
        Weight Decay                 & 0 \\
        AdamW $\beta_1, \beta_2$     & 0.9, 0.999 \\
        \midrule
        \multicolumn{2}{c}{QLoRA Finetuning for 70B models\tnote{b}} \\ 
        \midrule
        Training Epoch               & 2    \\ 
        LoRA Rank                    & 64   \\
        LoRA Alpha                   & 16   \\
        LoRA Dropout                 & 0.1  \\
        Learning Rate                & 1e-4 \\
        Warmup Ratio                 & 0.03 \\
        \midrule
        \multicolumn{2}{c}{Weak Supervision Experiment} \\ 
        \midrule
        $\lambda$ for binary classifier loss    & 0.5 \\
        Positive Class Weight in binary loss    & 8 or 5\tnote{c} \\
        \bottomrule
        \end{tabular}
        \small 
        \begin{tablenotes}
            \item[a] The settings are similar to the ones used by \citet{liu2024tuning}. 
            \item[b] We adopt the same values as \citet{ivison2023camels}.
            \item[c] 8 for \datasettulu training and 5 for the rest, set based on the class imbalance in the training data.
        \end{tablenotes}
    \end{threeparttable}
    }
\caption{Training hyperparameters for our experiments.}
\label{table:hyperparameters}
\end{table}

\begin{table}[h]
    \resizebox{1.\linewidth}{!}{
    \begin{threeparttable}
        \renewcommand{\arraystretch}{1.20}
        \setlength{\tabcolsep}{4pt}
        \begin{tabular}{lllrr}
        \toprule
        & \textbf{Base}       & \textbf{Asst.}       & \multicolumn{1}{c}{$\eta_*$} & \multicolumn{1}{c}{$f$} \\
        \midrule
        \begin{tabular}[c]{@{}l@{}}\datasetalpacaeval\\ $N=2048$\end{tabular}             & \llama[7B] & \llama[70B] & 0.80                         & 0.1                     \\
        \midrule
        \multirow{3}{*}{\begin{tabular}[c]{@{}l@{}}\datasetgsm\\ $N=512$\end{tabular}}    & \llama[7B] & \llama[70B] & 0.17                         & 0.1                     \\
        & \llama[7B] & \llm{Llemma}-7B    & 0.08 & 0.1 \\
        & \llama[7B] & \llm{Llemma}-34B   & 0.05 & 0.3 \\
        \cdashline{2-5}[.4pt/1pt]\noalign{\vskip 1pt}
        & \llm{Mistral}-7B & \llm{Mixtral}-8$\times$7B   & 0.12 & 0.2 \\
        \midrule
        \multirow{3}{*}{\begin{tabular}[c]{@{}l@{}}\datasetmath\\ $N=512$\end{tabular}}   & \llama[7B] & \llama[70B] & 0.57                         & 0.6                     \\
        & \llama[7B] & \llm{Llemma}-7B    & 0.05 & 0.9 \\
        & \llama[7B] & \llm{Llemma}-34B   & 0.30 & 0.8 \\
        \cdashline{2-5}[.4pt/1pt]\noalign{\vskip 1pt}
        & \llm{Mistral}-7B & \llm{Mixtral}-8$\times$7B   & 0.69 & 0.9 \\
        \midrule
        \multirow{3}{*}{\begin{tabular}[c]{@{}l@{}}\datasetbioasq\\ $N=512$\end{tabular}} & \llama[7B] & \llama[70B] & 0.38                         & 0.2                     \\
        & \llama[7B] & \llm{Meditron}-7B  & 0.07 & 0.5 \\
        & \llama[7B] & \llm{Meditron}-70B & 0.20 & 0.5 \\
        \bottomrule
        \end{tabular}
    \end{threeparttable}
    }
\caption{For each dataset, we show the max number of generated tokens $N$ (used for all models) and the optimal deferral threshold $\eta_*$ (corresponding frequency $f$) used to generate the responses (available only for \ourmethod).}
\label{table:evaluation_setup}
\vspace{-1mm}
\end{table}

\begin{table}[h]
    \resizebox{1.\linewidth}{!}{
    \begin{threeparttable}
        \renewcommand{\arraystretch}{1.20}
        \setlength{\tabcolsep}{4pt}
        \begin{tabular}{lllrr}
        \toprule
        & \textbf{Base}       & \textbf{Asst.}       & \multicolumn{1}{c}{$\eta_*$} & \multicolumn{1}{c}{$f$} \\
        \midrule
        \datasetalpacaeval & \llama[7B] & \llama[70B] & 0.11 & 0.1 \\
        \datasetgsm        & \llama[7B] & \llama[70B] & 1.00 & 0.0 \\
        \datasetmath       & \llama[7B] & \llama[70B] & 0.44 & 0.1 \\
        \datasetbioasq     & \llama[7B] & \llama[70B] & 1.00 & 0.0 \\
        \bottomrule
        \end{tabular}
    \end{threeparttable}
    }
\caption{Similar to \cref{table:evaluation_setup}, we report the the optimal deferral threshold $\eta_*$ and frequency used for Weakly-supervised \ourmethod.}
\label{table:evaluation_setup_weak_supervision}

\end{table}

\begin{table}[!h]
    \centering
    \small
    \begin{threeparttable}
        \renewcommand{\arraystretch}{1.15}
        \setlength{\tabcolsep}{4pt}
        \begin{tabular}{lrrr}
        \toprule
            & \multicolumn{3}{c}{\# of Samples} \\
        \textbf{Dataset}            & \textbf{Train}      & \textbf{Dev}      & \textbf{Test}     \\
        \midrule
        \datasettulu       & 326,115      &          &          \\
        \datasetmtbench    &            & 24       &          \\
        \datasetalpacaeval &            &          & 805      \\
        \midrule
        \datasetgsm        & 7,473       & 50       & 200      \\
        \datasetmath       & 7,498       & 60       & 750      \\
        \datasetbioasq     & 4,719       & 20       & 310      \\
        \cdashline{1-4}[.4pt/1pt]\noalign{\vskip 1pt}
        \multicolumn{3}{r}{\small (Factiod)}  & {\small 93}  \\
        \multicolumn{3}{r}{\small (List)}  &  {\small 61}  \\
        \multicolumn{3}{r}{\small (Summary)} &  {\small 75}  \\
        \multicolumn{3}{r}{\small (Yes/No) }&  {\small 81} \\
        \bottomrule
        \end{tabular}
    \end{threeparttable}
\caption{The training and evaluation dataset sizes. For the instruction following task, we train a model on the T\"ulu mixture, uses a small validation set from the \datasetmtbench dataset to pick the deferral threshold $\eta$, and evaluate on the \datasetalpacaeval dataset. For the other tasks, we train and test in-domain.}
\label{table:dataset}
\vspace{-3mm}
\end{table}

\begin{table}[!h]
    \small
    \centering
    \begin{threeparttable}
        \renewcommand{\arraystretch}{1.20}
        \setlength{\tabcolsep}{8pt}
        \begin{tabular}{lr}
        \toprule
        \textbf{Dataset Name} & \textbf{License} \\ 
        \midrule
        \datasetalpacaeval & CC BY-NC 4.0 \\
        \datasetgsm        & MIT \\ 
        \datasetmath       & MIT \\
        \datasetbioasq     & CC BY 2.5 \\
        T\"ulu v2 mix      & ODC BY \\
        \bottomrule
        \end{tabular}
    \end{threeparttable}
\caption{The licenses of the datasets used.}
\label{table:dataset-licenses}
\end{table}

\begin{algorithm}[!h]
    \SetAlgoLined
    \SetKwInOut{Input}{Input}
    \SetKw{KwBy}{by}
    \Input{Base Model, Asst. Model, Model Selector $\phi$, Validation Dataset $\mathcal{D}$}
    Let $\mathcal{P} = \{\}$ \\
    \For{$i=0$ \KwTo $|\mathcal{D}|$}{
        Given the input prompt in $\mathcal{D}_i$, generate response $X^{(i)}$ using the base model \\
        Predict per-token deferral probability {\small $P_\phi(Z_t^{(i)} =1 \mid X_{<t}^{(i)})$}, and append it to $\mathcal{P}$ \\
    }
    Sort $\mathcal{P}$ in ascending order \\
    Set current best threshold $\eta = 0$ and evaluation score $s=0$\\
    \For{$j=0$ \KwTo $100$ \KwBy $10$}{
        Get the $j$-th quantile $p_j$ in $\mathcal{P}$ and use it as the deferral threshold $\eta_j$ \\ 
        Generate responses $X^{(i)}$ for $i$ in from $0$ to $|\mathcal{D}|$ using the base and asst. model controlled by {\small $P_\phi(Z_t^{(i)}=1 \mid X_{<t}^{(i)})$ $>\eta_j$} \\
        Evaluate the responses, and if the evaluation score is better than $s$, set $\eta = \eta_j$ and $s$ to the new score \\
    }
    Return $\eta$ \\
    \caption{Find Optimal Deferral Th. $\eta$}
    \label{alg:search-deferral-threshold}
\end{algorithm}

\subsection{\datasetbioasq Evaluation} 

There are four different substasks in the \datasetbioasq dataset, and they are evaluated using different metrics according to the guideline:\footnote{\url{http://participants-area.bioasq.org/general_information/Task9b/}}

\tightlist{
    \item \textbf{Factoid}: It require a particular entity name (e.g., of a disease, drug, or gene, or a number) to be generated per the question, which might not appera in the original question. The model may generate a list of candidate, and we pick the first generation (as the model often only generates one) and search for matching among the allowed candidate answers. This is the Strict Accuracy (SAcc.) metric in~\citet{krithara2023bioasq}. 
    \item \textbf{List}: Similar to Factoid questions but the model is expected to generate a list of entities. The model is required to produce the answer in a bullet list format, and we use F1 score to evaluate the performance.
    \item \textbf{Summary}: The answer is expected to be a long-form text like the description of a treatment. Following \citet{krithara2023bioasq}, we report ROUGE-2~\citep{lin-2004-rouge} (using the implementation from the Huggingface Evaluate library\footnote{\url{https://huggingface.co/spaces/evaluate-metric/rouge}}) to measure the textual overlapping between the generation and the ground truth.
    \item \textbf{Yes/No}: The model needs to provide a binary answer yes or no to the given question, and the classification accuracy is reported. 
}

\section{Additional Generation Examples}
\cref{fig:decoding-example} shows an simplified version of generation for clarity, and we show the original generation in \cref{table:method_figure_example}. 
An additional example generation on \datasetmath is shown in \cref{table:exampe-5}.

\begin{table}[!h]
    \small 
    \centering
    \renewcommand{\arraystretch}{1.20}
    \begin{tabular}{p{0.9\linewidth}}
        \toprule
        \texttt{Op}\texttt{du}\defer{\texttt{al}}\texttt{ag}\texttt{ contains}\defer{\texttt{ two}}\defer{\texttt{ active}}\texttt{ components}\texttt{:}\texttt{ }\defer{\texttt{1}}\texttt{)}\defer{\texttt{ n}}\defer{\texttt{iv}}\defer{\texttt{ol}}\defer{\texttt{um}}\texttt{ab}\texttt{ and}\texttt{ }\defer{\texttt{2}}\texttt{)}\texttt{ rel}\defer{\texttt{at}}\defer{\texttt{lim}}\defer{\texttt{ab}}\texttt{.}\texttt{
}\texttt{The}\texttt{ final}\texttt{ answer}\texttt{ is}\texttt{:}\texttt{ }\defer{\texttt{1}}\texttt{)}\texttt{ n}\texttt{iv}\texttt{ol}\texttt{um}\texttt{ab}\texttt{ }\defer{\texttt{2}}\texttt{)}\texttt{ rel}\texttt{at}\texttt{lim}\texttt{ab}
        \\
        \bottomrule
    \end{tabular}
    \caption{The original (token-level) model generation for the example in \cref{fig:decoding-example}. \cref{fig:decoding-example} shows the generation at the word level rather than the token level for simplicity. }
    \label{table:method_figure_example}
    \vspace{5mm}
\end{table}

\begin{table*}[t]
    \centering
    \begin{threeparttable}
        \renewcommand{\arraystretch}{1.20}
        \setlength{\tabcolsep}{4pt}
        \begin{tabular}{lm{0.8\linewidth}}
        \toprule
        \textbf{Dataset} & \textbf{Prompt} \\
        \midrule
        \datasetalpacaeval & \texttt{<|user|>}\newline\promptData{What are some species of bears that are now extinct?}\newline\texttt{<|assistant|>}\newline \\
        \midrule
        \datasetgsm & \texttt{Please solve the following math problem with detailed steps.}
        \newline \newline \texttt{Question:} \promptData{Tom's ship can travel at 10 miles per hour. He is sailing from 1 to 4 PM. He then travels back at a rate of 6 mph. How long does it take him to get back?} \newline \texttt{Answer:}\\
        \midrule
        \datasetmath & \texttt{Given a mathematics problem, determine the answer. Simplify your answer as much as possible. You can use latex to format your answer and you should state your final answer as ``The final answer is \$(final-answer)\$.''} \newline \newline \texttt{Problem:} \promptData{Consider the function \$g(x)=3x-4\$.  What is \$g(0)\$?} \newline \newline \texttt{Solution:}\\
        \bottomrule
        \datasetbioasq & \texttt{You are a medical expert answering real-world biomedical questions. Based on your understanding of basic and clinical science, medical knowledge, and mechanisms underlying health, disease, patient care, and modes of therapy, write a list of entity names, numbers, or short expressions that answer the following question.}
        \newline \newline 
        \texttt{Question:} \promptData{Which genes are related to Emery-Dreifuss Muscular Dystrophy (EDMD)?}
        \newline
        \texttt{Answer:}
        \\
        \bottomrule
        \end{tabular}
    \end{threeparttable}
\caption{Prompts used for formatting the datasets during training and evaluation. For training on the T\"ulu mixture, we adopt the same chat template as~\citet{Wang2023HowFC}. The prompt for \datasetgsm and \datasetmath is based on the prompt by \citet{wei2022chain} and ~\citet{hendrycks2021measuring}. We modify the prompt for \texttt{MedMCQA} questions in the Meditron paper~\cite{chen2023meditron70b} for \datasetbioasq, and adapt it for different tasks.}
\label{table:prompts}
\end{table*}

\begin{table*}[t]
    \resizebox{1.\linewidth}{!}{
    \begin{tabular}{p{0.05\textwidth}p{0.5\textwidth}p{0.45\textwidth}}
    \toprule
    $f$ & \textbf{Generation (Source \TeX)} & \textbf{Rendered} \\
    \midrule
    GT 
    & 
    \texttt{Comple}\texttt{ting}\texttt{ the}\texttt{ square}\texttt{ in}\texttt{ \$}\texttt{x}\texttt{\$}\texttt{ and}\texttt{ \$}\texttt{y}\texttt{,\$}\texttt{ we}\texttt{ get}\texttt{ }\texttt{\textbackslash{}}\texttt{[}\texttt{9}\texttt{(}\texttt{x}\texttt{ +}\texttt{ }\texttt{4}\texttt{)\textasciicircum{}}\texttt{2}\texttt{ +}\texttt{ }\texttt{4}\texttt{(}\texttt{y}\texttt{ -}\texttt{ }\texttt{1}\texttt{)\textasciicircum{}}\texttt{2}\texttt{ =}\texttt{ }\texttt{3}\texttt{2}\texttt{4}\texttt{.\textbackslash{}}\texttt{]}
    \texttt{Then}\texttt{ }\texttt{\textbackslash{}}\texttt{[\textbackslash{}}\texttt{frac}\texttt{\{(}\texttt{x}\texttt{ +}\texttt{ }\texttt{4}\texttt{)\textasciicircum{}}\texttt{2}\texttt{\}\{}\texttt{3}\texttt{6}\texttt{\}}\texttt{ +}\texttt{ \textbackslash{}}\texttt{frac}\texttt{\{(}\texttt{y}\texttt{ -}\texttt{ }\texttt{1}\texttt{)\textasciicircum{}}\texttt{2}\texttt{\}\{}\texttt{8}\texttt{1}\texttt{\}}\texttt{ =}\texttt{ }\texttt{1}\texttt{.\textbackslash{}}\texttt{]}\texttt{Th}\texttt{us}\texttt{,}\texttt{ the}\texttt{ center}\texttt{ of}\texttt{ the}\texttt{ el}\texttt{lipse}\texttt{ is}\texttt{ \$\textbackslash{}}\texttt{box}\texttt{ed}\texttt{\{}\texttt{(-}\texttt{4}\texttt{,}\texttt{1}\texttt{)\}.}\texttt{\$}
    & 
    {\footnotesize Completing the square in $x$ and $y,$ we get \[9(x + 4)^2 + 4(y - 1)^2 = 324.\]Then \[\frac{(x + 4)^2}{36} + \frac{(y - 1)^2}{81} = 1.\]Thus, the center of the ellipse is $\boxed{(-4,1)}.$} \\
    \midrule
    0.0 & 
    \texttt{Comple}\texttt{ting}\texttt{ the}\texttt{ square}\texttt{ in}\texttt{ \$}\texttt{x}\texttt{\$}\texttt{ and}\texttt{ \$}\texttt{y}\texttt{,\$}\texttt{ we}\texttt{ get}\texttt{ \textbackslash{}[}\texttt{9}\texttt{x}\texttt{\textasciicircum{}}\texttt{2}\texttt{ +}\texttt{ }\texttt{7}\texttt{2}\texttt{x}\texttt{ +}\texttt{ }\texttt{4}\texttt{y}\texttt{\textasciicircum{}}\texttt{2}\texttt{ -}\texttt{ }\texttt{8}\texttt{y}\texttt{ -}\texttt{ }\texttt{1}\texttt{7}\texttt{6}\texttt{ =}\texttt{ }\texttt{0}\texttt{ \textbackslash{}}\texttt{Rightarrow}\texttt{ }\texttt{9}\texttt{x}\texttt{\textasciicircum{}}\texttt{2}\texttt{ +}\texttt{ }\texttt{7}\texttt{2}\texttt{x}\texttt{ +}\texttt{ }\texttt{4}\texttt{y}\texttt{\textasciicircum{}}\texttt{2}\texttt{ -}\texttt{ }\texttt{8}\texttt{y}\texttt{ -}\texttt{ }\texttt{1}\texttt{7}\texttt{6}\texttt{ =}\texttt{ }\texttt{9}\texttt{(}\texttt{x}\texttt{ -}\texttt{ }\texttt{2}\texttt{)\textasciicircum{}}\texttt{2}\texttt{ +}\texttt{ }\texttt{7}\texttt{2}\texttt{(}\texttt{x}\texttt{ -}\texttt{ }\texttt{2}\texttt{)}\texttt{ +}\texttt{ }\texttt{4}\texttt{(}\texttt{y}\texttt{ -}\texttt{ }\texttt{2}\texttt{)\textasciicircum{}}\texttt{2}\texttt{ -}\texttt{ }\texttt{8}\texttt{(}\texttt{y}\texttt{ -}\texttt{ }\texttt{2}\texttt{)}\texttt{ -}\texttt{ }\texttt{1}\texttt{7}\texttt{6}\texttt{.\textbackslash{}}\texttt{]}\texttt{Then}\texttt{ \textbackslash{}[\textbackslash{}}\texttt{box}\texttt{ed}\texttt{\{}\texttt{2}\texttt{\}}\texttt{ =}\texttt{ \textbackslash{}}\texttt{frac}\texttt{\{}\texttt{1}\texttt{\}\{}\texttt{9}\texttt{\}(}\texttt{9}\texttt{(}\texttt{x}\texttt{ -}\texttt{ }\texttt{2}\texttt{)\textasciicircum{}}\texttt{2}\texttt{ +}\texttt{ }\texttt{7}\texttt{2}\texttt{(}\texttt{x}\texttt{ -}\texttt{ }\texttt{2}\texttt{)}\texttt{ +}\texttt{ }\texttt{4}\texttt{(}\texttt{y}\texttt{ -}\texttt{ }\texttt{2}\texttt{)\textasciicircum{}}\texttt{2}\texttt{ -}\texttt{ }\texttt{8}\texttt{(}\texttt{y}\texttt{ -}\texttt{ }\texttt{2}\texttt{)}\texttt{ -}\texttt{ }\texttt{1}\texttt{7}\texttt{6}\texttt{).}\texttt{\textbackslash{}]}\texttt{ The}\texttt{ final}\texttt{ answer}\texttt{ is}\texttt{ \$}\texttt{2}\texttt{\$.}
    & 
    {\footnotesize Completing the square in $x$ and $y,$ we get$$\begin{aligned} & 9x^2 + 72x + 4y^2 - 8y - 176 = 0 \Rightarrow \\ &  9x^2 + 72x + 4y^2 - 8y - 176 =\\ &  9(x - 2)^2 + 72(x - 2) + 4(y - 2)^2 \\ &  - 8(y - 2) - 176.\end{aligned}$$Then $$\begin{aligned} & \boxed{2} = \frac{1}{9}(9(x - 2)^2 + 72(x - 2) \\ & + 4(y - 2)^2 - 8(y - 2) - 176).\end{aligned}$$ The final answer is $2$.}
    \\ \cdashline{1-3}[.4pt/1pt]\noalign{\vskip 1pt}
    0.4 &  
    \texttt{Comple}\texttt{ting}\texttt{ the}\texttt{ square}\texttt{ in}\texttt{ \$}\texttt{x}\texttt{\$}\texttt{ and}\texttt{ \$}\texttt{y}\texttt{,\$}\texttt{ we}\texttt{ get}\texttt{ \textbackslash{}[}\texttt{9}\texttt{x}\texttt{\textasciicircum{}}\texttt{2}\texttt{ +}\texttt{ }\defer{\texttt{7}}\texttt{2}\texttt{x}\texttt{ +}\texttt{ }\defer{\texttt{4}}\texttt{y}\texttt{\textasciicircum{}}\texttt{2}\texttt{ -}\texttt{ }\defer{\texttt{8}}\texttt{y}\texttt{ -}\texttt{ }\texttt{1}\texttt{7}\texttt{6}\texttt{ =}\texttt{ }\defer{\texttt{0}}\texttt{ \textbackslash{}}\texttt{Rightarrow}\texttt{ }\texttt{9}\defer{\texttt{(}}\defer{\texttt{x}}\texttt{ -}\defer{\texttt{ }}\defer{\texttt{4}}\texttt{)\textasciicircum{}}\texttt{2}\texttt{ +}\defer{\texttt{ }}\defer{\texttt{4}}\defer{\texttt{(}}\defer{\texttt{y}}\texttt{ -}\defer{\texttt{ }}\defer{\texttt{1}}\defer{\texttt{)\textasciicircum{}}}\texttt{2}\defer{\texttt{ =}}\defer{\texttt{ }}\defer{\texttt{1}}\texttt{.\textbackslash{}}\texttt{]}\texttt{Then}\texttt{ \textbackslash{}[\textbackslash{}}\defer{\texttt{frac}}\texttt{\{}\defer{\texttt{9}}\defer{\texttt{(}}\defer{\texttt{x}}\texttt{ -}\texttt{ }\defer{\texttt{4}}\defer{\texttt{)\textasciicircum{}}}\defer{\texttt{2}}\defer{\texttt{\}\{}}\defer{\texttt{1}}\defer{\texttt{\}}}\texttt{ +}\defer{\texttt{ \textbackslash{}}}\defer{\texttt{frac}}\texttt{\{}\defer{\texttt{4}}\texttt{(}\defer{\texttt{y}}\texttt{ -}\texttt{ }\texttt{1}\texttt{)\textasciicircum{}}\texttt{2}\texttt{\}\{}\defer{\texttt{1}}\texttt{\}}\texttt{ =}\defer{\texttt{ }}\defer{\texttt{1}}\texttt{.\textbackslash{}}\texttt{]}\texttt{H}\texttt{ence}\texttt{,}\texttt{ \textbackslash{}[}\defer{\texttt{a}}\defer{\texttt{ =}}\defer{\texttt{ \textbackslash{}}}\defer{\texttt{sqrt}}\defer{\texttt{\{}}\defer{\texttt{9}}\defer{\texttt{\}}}\defer{\texttt{ =}}\defer{\texttt{ }}\defer{\texttt{3}}\texttt{,\textbackslash{}}\texttt{]}\texttt{and}\texttt{ \$}\texttt{b}\texttt{ =}\defer{\texttt{ \textbackslash{}}}\defer{\texttt{sqrt}}\defer{\texttt{\{}}\defer{\texttt{4}}\texttt{\}}\texttt{ =}\defer{\texttt{ }}\defer{\texttt{2}}\texttt{.\$}\texttt{ }\texttt{ Then}\texttt{ the}\texttt{ center}\texttt{ of}\texttt{ the}\texttt{ el}\texttt{lipse}\texttt{ is}\texttt{ \$\textbackslash{}}\texttt{box}\texttt{ed}\defer{\texttt{\{\textbackslash{}}}\defer{\texttt{textbf}}\texttt{\{}\defer{\texttt{ (}}\texttt{4}\texttt{,}\defer{\texttt{ }}\defer{\texttt{1}}\texttt{)\}\}}\texttt{.\$}\texttt{ The}\texttt{ final}\texttt{ answer}\texttt{ is}\texttt{ \$\textbackslash{}}\texttt{textbf}\texttt{\{}\texttt{ (}\texttt{4}\texttt{,}\texttt{ }\texttt{1}\texttt{)}\texttt{\}\$.}
    &  
    {\footnotesize   Completing the square in $x$ and $y,$ we get $$\begin{aligned} & 9x^2 + \defereq{7}2x + \defereq{4}y^2 - \defereq{8}y - 176 = \defereq{0} \Rightarrow \\ & 9\defereq{(x} -\defereq{ 4})^2 + \defereq{4(y} - \defereq{1)}^2 \defereq{= 1}.\end{aligned}$$ Then \[\frac{\defereq{9(x} - \defereq{4)^2}}{\defereq{1}} + \frac{\defereq{4(y} - 1)^2}{\defereq{1}} = \defereq{1}.\]Hence, \[\defereq{a = \sqrt{9} = 3},\]and $b = \defereq{\sqrt{4} = 2}.$  Then the center of the ellipse is $\boxed{\textbf{ \defereq{(}4, \defereq{ 1})}}.$ The final answer is $\textbf{ (4, 1)}$.}
    \\ \cdashline{1-3}[.4pt/1pt]\noalign{\vskip 1pt}
    0.8 &
    \texttt{Comple}\texttt{ting}\texttt{ the}\texttt{ square}\texttt{ in}\texttt{ \$}\texttt{x}\texttt{\$}\texttt{ and}\texttt{ \$}\texttt{y}\texttt{,\$}\texttt{ we}\texttt{ get}\texttt{ \textbackslash{}[}\defer{\texttt{9}}\defer{\texttt{(}}\defer{\texttt{x}}\defer{\texttt{\textasciicircum{}}}\defer{\texttt{2}}\defer{\texttt{ +}}\defer{\texttt{ }}\defer{\texttt{8}}\defer{\texttt{x}}\defer{\texttt{)}}\defer{\texttt{ +}}\defer{\texttt{ }}\defer{\texttt{4}}\defer{\texttt{(}}\defer{\texttt{y}}\defer{\texttt{\textasciicircum{}}}\defer{\texttt{2}}\defer{\texttt{ -}}\defer{\texttt{ }}\defer{\texttt{2}}\defer{\texttt{y}}\defer{\texttt{)}}\defer{\texttt{ =}}\defer{\texttt{ }}\defer{\texttt{1}}\defer{\texttt{7}}\defer{\texttt{6}}\texttt{.\textbackslash{}}\texttt{]}\texttt{Then}\texttt{ \textbackslash{}[\textbackslash{}}\defer{\texttt{begin}}\texttt{\{}\texttt{aligned}\texttt{\}}\defer{\texttt{ }}\defer{\texttt{9}}\defer{\texttt{(}}\defer{\texttt{x}}\defer{\texttt{\textasciicircum{}}}\defer{\texttt{2}}\defer{\texttt{ +}}\defer{\texttt{ }}\defer{\texttt{8}}\defer{\texttt{x}}\defer{\texttt{ +}}\defer{\texttt{ }}\defer{\texttt{1}}\defer{\texttt{6}}\defer{\texttt{)}}\defer{\texttt{ +}}\defer{\texttt{ }}\defer{\texttt{4}}\defer{\texttt{(}}\defer{\texttt{y}}\defer{\texttt{\textasciicircum{}}}\defer{\texttt{2}}\defer{\texttt{ -}}\defer{\texttt{ }}\defer{\texttt{2}}\defer{\texttt{y}}\defer{\texttt{ +}}\defer{\texttt{ }}\defer{\texttt{1}}\defer{\texttt{)}}\defer{\texttt{ \&=}}\defer{\texttt{ }}\defer{\texttt{1}}\defer{\texttt{7}}\defer{\texttt{6}}\texttt{ \textbackslash{}\textbackslash{}}\defer{\texttt{ }}\defer{\texttt{9}}\defer{\texttt{(}}\defer{\texttt{x}}\defer{\texttt{ +}}\defer{\texttt{ }}\defer{\texttt{4}}\defer{\texttt{)\textasciicircum{}}}\defer{\texttt{2}}\defer{\texttt{ +}}\defer{\texttt{ }}\defer{\texttt{4}}\defer{\texttt{(}}\defer{\texttt{y}}\defer{\texttt{ -}}\defer{\texttt{ }}\defer{\texttt{1}}\defer{\texttt{)\textasciicircum{}}}\defer{\texttt{2}}\defer{\texttt{ \&=}}\defer{\texttt{ }}\defer{\texttt{1}}\defer{\texttt{7}}\defer{\texttt{6}}\texttt{.}\texttt{ \textbackslash{}}\texttt{end}\defer{\texttt{\{}}\defer{\texttt{aligned}}\texttt{\}\textbackslash{}}\texttt{]}\texttt{There}\defer{\texttt{fore}}\texttt{,}\defer{\texttt{ the}}\defer{\texttt{ center}}\texttt{ of}\defer{\texttt{ the}}\texttt{ el}\texttt{lipse}\texttt{ is}\texttt{ \$\textbackslash{}}\texttt{box}\texttt{ed}\defer{\texttt{\{}}\defer{\texttt{(-}}\defer{\texttt{4}}\defer{\texttt{,}}\defer{\texttt{ }}\defer{\texttt{1}}\defer{\texttt{)\}.}}\texttt{\$}\texttt{ The}\texttt{ final}\texttt{ answer}\texttt{ is}\texttt{ \$(}\texttt{-}\texttt{4}\texttt{,}\texttt{ }\texttt{1}\texttt{)\$.}
    &
    {\footnotesize  Completing the square in $x$ and $y,$ we get \[\defereq{9(x^2 + 8x) + 4(y^2 - 2y) = 176.}\]Then \[\begin{aligned} \defereq{9(x^2 + 8x + 16) + 4(y^2 - 2y + 1)} &\defereq{= 176} \\ \defereq{9(x + 4)^2 + 4(y - 1)^2} &\defereq{= 176}. \end{aligned}\]There\defer{fore}, \defer{the center} of \defer{the} ellipse is $\boxed{\defereq{(-4, 1)}}.$ The final answer is $(-4, 1)$.} 
    \\ \bottomrule 
    \end{tabular}
    }
    \caption{Model answers with different deferral frequencies to the question ``Find the center of the ellipse whose equation is $9x^2 + 72x + 4y^2 - 8y - 176 = 0.$'' In this example, the base model is a \llama[7B] model fine-tuned on the MATH dataset, and the reference model is the \llm{Llemma}-34B model. We show the ground-truth answer in the first row, and the 0-shot model answers with different deferral ratios, with the original outputs and rendered. We use yellow background to indicate the token is produced by the reference model rather than the base model.}
    \label{table:exampe-5}
\end{table*}

\end{document}